\documentclass[journal,,10pt,onecolumn]{IEEEtran}
\usepackage{cite}
\usepackage{mathptm}    
\usepackage{mathptmx} 
\usepackage{mathbbol} 
\usepackage{amsmath} 
\usepackage{stfloats}
\usepackage{bm}
\usepackage{multirow}
\usepackage{pslatex}	
\usepackage{booktabs}
\usepackage[utf8]{inputenc}   
\usepackage{amssymb}
\usepackage{amsthm}
\usepackage{amsmath}
\usepackage[ruled,lined,linesnumbered]{algorithm2e}
\usepackage{graphicx}
\usepackage{epstopdf}
\usepackage[normalem]{ulem}
\usepackage{array}
\usepackage{makecell}
\usepackage{color}

\DeclareUnicodeCharacter{2061}{}

\providecommand{\keywords}[1]{\textbf{\textit{Index terms---}} #1}

\begin{document}
	
"This work has been submitted to the IEEE for possible publication. Copyright may be transferred without notice, after which this version may no longer be accessible."
\pagebreak
	
	\title{ Joint, Partially-joint, and Individual Independent Component Analysis in Multi-Subject fMRI Data}
	
	\author{Mansooreh Pakravan and Mohammad Bagher Shamsollahi,~\textit{Senior Member, IEEE}
		\thanks{M. Pakravan and M.B, Shamsolahi are with the Department of Electrical Engineering, Sharif 
			University of Technology, Tehran, Iran (e-mail: mpakravan@ee.sharif.edu, mbshams@sharif.edu). }}

	\maketitle
	
	\begin{abstract}
		\textit{Objective}: Joint analysis of multi-subject brain imaging datasets has wide applications in biomedical engineering. In these datasets, some sources belong to all subjects (joint), a subset of subjects (partially-joint), or a single subject (individual). In this paper, this source model is referred to as joint/partially-joint/individual multiple datasets multidimensional (JpJI-MDM), and accordingly, a source extraction method is developed. \textit{Method}: We present a deflation-based algorithm utilizing higher order cumulants to analyze the JpJI-MDM source model. The algorithm maximizes a cost function which leads to an eigenvalue problem solved with \textcolor[rgb]{0,0,0}{thin-SVD (singular value decomposition) factorization}. Furthermore, we introduce the\textit{ JpJI-feature} which indicates the spatial shape of each source and the amount of its jointness with other subjects. We use this feature to determine the type of sources. \textit{Results}: We evaluate our algorithm by analyzing simulated data and two real functional magnetic resonance imaging (fMRI) datasets. In our simulation study, we will show that the proposed algorithm determines the type of sources with the accuracy of 95\% and 100\% for 2-class and 3-class clustering scenarios, respectively. Furthermore, our algorithm extracts meaningful joint and partially-joint sources from the two real datasets, which are consistent with the existing neuroscience studies. \textit{Conclusion}: Our results in analyzing the real datasets reveal that both datasets follow the JpJI-MDM source model. This source model improves the accuracy of source extraction methods developed for multi-subject datasets. \textit{Significance}: The proposed joint blind source separation algorithm is robust and avoids parameters which are difficult to fine-tune. 
	\end{abstract}
	
	\keywords{Joint analysis, Multi-subject dataset, Partially-joint sources, Independent component analysis, Multiple dataset unidimensional }

	\section{Introduction} 
	\subsection {Background and Motivation}
	Integrating information of multi-subject datasets is receiving much attention in biomedical engineering and neuroscience. With the joint analysis of multiple datasets and the fusion of their data, the cross-information of datasets is extracted. The benefit of joint analysis is more significant for datasets with common information, thereby the accuracy and validity of extracted sources are improved. Employing proper methods to extract relevant information from multiple datasets is of paramount importance because many unknown variables affect the underlying processes. Blind source separation (BSS) methods are useful data-driven techniques to jointly analyze multiple datasets \cite{ref1}.

	The general BSS sub-problems can be categorized into four classes \cite{ref4}, depending on the number of analyzed datasets ($K$) and the dependency model of sources. These classes are single dataset unidimensional (SDU), multiple datasets unidimensional (MDU), single dataset multi-dimensional (SDM), and multiple datasets multi-dimensional (MDM). In Table I, the features and related methods for each class are summarized. A special case for MDM source model is joint/individual MDM (JI-MDM) source model \cite{ref30}\cite{pakravan2018}, in which each dataset has two parts, namely joint sources and individual sources, where joint sources have exactly one equivalent (similar) source across all datasets, while individual sources are independent of other sources in that dataset and other datasets.

	\begin{table*}[]
		\caption{Source models used in BSS subproblems}
		\resizebox{\textwidth}{!}{%
			\label{Table0}
			\begin{tabular}{|c|l|l|}
				\hline
				\textbf{\begin{tabular}[c]{@{}c@{}}General BSS \\ sub-problems\end{tabular}}      & \multicolumn{1}{c|}{\textbf{Features}}                                                                                                                                                                & \multicolumn{1}{c|}{\textbf{Related methods}}                                                                                                                                                                                                                                                                                                                                                                                     \\ \hline
				\begin{tabular}[c]{@{}c@{}}Single dataset \\ unidimensional\\ (SDU)\end{tabular}     & \begin{tabular}[c]{@{}l@{}}-$K=1$\\ -Uncorrelated or independent sources\end{tabular}                                                                                                                 & \begin{tabular}[c]{@{}l@{}}-Principal component analysis (PCA) \cite{ref1}\\ -Independent component analysis (ICA) \cite{ref1}\\ -Second order blind identification (SOBI) methods \cite{ref1} \end{tabular}                                                                                                                                                                            \\ \hline
				\begin{tabular}[c]{@{}c@{}}Multiple dataset\\ unidimensional \\ (MDU)\end{tabular}   & \begin{tabular}[c]{@{}l@{}}-$K>1$\\ -Independent sources in each dataset \\ -Exactly one dependent source\\  between datasets\end{tabular}                                                            & \begin{tabular}[c]{@{}l@{}}-Canonical correlation analysis (CCA) \cite{ref11}\\ -Common feature analysis method \cite{zhang2015ssvep}\\ -Cross cumulant tensor block diagonalization \cite{ref12}\\ -Group information guided ICA (GIG-ICA) \cite{ref40}\\ -Independent vector analysis (IVA) \cite{ref25}\end{tabular} \\ \hline
				\begin{tabular}[c]{@{}c@{}}Single dataset \\ multi-dimensional\\ (SDM)\end{tabular}   & \begin{tabular}[c]{@{}l@{}}-$K=1$\\ -One or more group of sources\\ -Dependent sources in each group\end{tabular}                                                                                     & \begin{tabular}[c]{@{}l@{}}-Multidimensional independent component analysis (MICA) \cite{ref13}\\ -Independent subspace analysis (ISA) \cite{ref14}\end{tabular}                                                                                                                                                                                                                                        \\ \hline
				\begin{tabular}[c]{@{}c@{}}Multiple dataset \\ multi-dimensional\\ (MDM)\end{tabular} & \begin{tabular}[c]{@{}l@{}}-$K>1$\\ -One or more group of sources in each dataset\\ -Dependent sources in each group\\ -Joint group of sources across all, some or\\ just one of datasets\end{tabular} & \begin{tabular}[c]{@{}l@{}}-Joint and individual variation explained (JIVE) \cite{ref26}\\ -Common orthogonal basis extraction (COBE) and common \\ nonnegative features extraction (CNFE) \cite{ref30}\\ -Joint/Individual Thin ICA (JI-ThICA) \cite{pakravan2018} \\ -Multi-dataset independent subspace analysis (MISA) \cite{ref15}\\ -Joint independent subspace analysis (JISA) \cite{ref16}\end{tabular}                    \\ \hline
			\end{tabular}}
		\end{table*}
		
		In \cite{pakravan2018}, it has been shown that the JI-MDM source model is suitable for multi-subject brain datasets, because real brain datasets, e.g., fMRI signals of multiple subjects in the same experiment, are not exactly similar in all regions across subjects. Clearly, there are some active processes in the brain of each subject that do not depend on the effect of the experiment. Thus, to enhance the validity of the source model of datasets, a number of individual (independent) sources should be considered for each dataset. In the JI-MDM source model, joint sources indicate potential relationships and cross-information among subjects, while individual sources indicate unique information of each subject. 
		
		In \cite{pakravan2018},  an algorithm, referred to as Joint/Individual thin independent component analysis (JI-ThICA), has been proposed to extract joint and individual sources of multi-subject datasets based on the JI-MDM source model. It is worth mentioning that the JI-ThICA algorithm has been inspired by the thin ICA algorithm \cite{ref2} \cite{cruces2004thin}, where the cost function of thin ICA is a proxy non-Gaussianity measure obtained by combining higher order cumulant matrices.
		
		In order to further improve the accuracy of the source model for multi-subject brain datasets, in this paper, we introduce a new source model in which three types of sources are considered, joint, partially-joint, and individual sources. Here joint and individual sources have the same meaning as the JI-MDM model; however, the newly defined partially-joint sources indicate sources that are common only among subsets of subjects. Hereafter, this source model is referred to as joint/partially-joint/individual MDM (JpJI-MDM). 
		
		The JpJI-MDM source model has wide applications in many studies analyzing multi-subject recordings from healthy and disease subjects. In these studies, all subjects may have some joint sources indicating common conditions of all subjects, some partially-joint sources which depend only on the conditions of the group of disease subjects (or healthy subjects), and some individual sources which are appeared due to the independent conditions of each subject. Thus, designing a new algorithm to analyze multi-subject datasets based on the JpJI-MDM source model is of utmost importance which is addressed in this paper. 
		
		There are lots of papers applying BSS techniques to analyze biomedical datasets. In Table I, the related papers are listed \cite{ref1, ref11, zhang2015ssvep, ref12, ref40, ref25, ref13, ref14, ref26,ref30,pakravan2018,ref15,ref16}. Among the methods proposed in these papers, JIVE \cite{ref26}, COBE \cite{ref30}, CNFE \cite{ref30}, and JI-ThICA \cite{pakravan2018}, which have the JI-MDM source model, are more relevant to this paper, however, none of them can extract partially-joint source.

		\subsection{Contribution and Paper Organization}
		
		To the best of our knowledge, there is no algorithm to jointly extract joint, partially-joint, and individual sources of multi-subject datasets. In this paper, we are motivated to address this issue and present a new data-driven algorithm utilizing higher order cumulants to analyze the JpJI-MDM source model. The algorithm maximizes a cost function to extract three types of sources. The maximization problem leads to an eigenvalue problem solved with \textcolor[rgb]{0,0,0}{thin-SVD factorization \cite{cruces2004thin}}. Furthermore, we introduce the \textit{JpJI-feature} which indicates the spatial shape of each source and the amount of its jointness with other datasets. We use this feature to determine the type of sources.
		
		\begin{figure}[t!]
			\centering
			\includegraphics[width=3.8in]{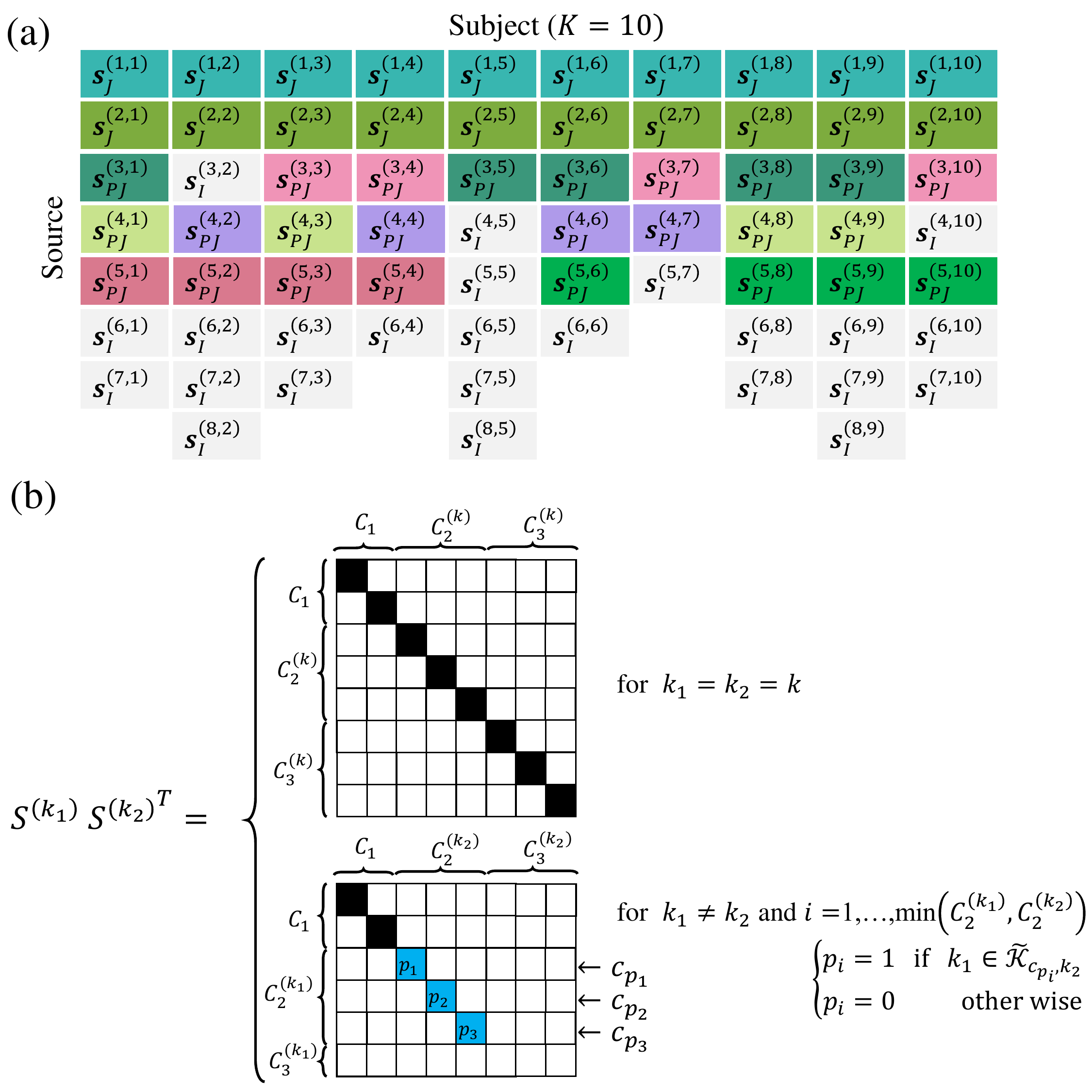}
			\caption{An example for JpJI-MDM source model. (a) Subjects with joint, partially-joint and individual sources such that $C_1=2$, $C_2=[3,2,3,3,1,3,2,3,3,2]$, and $C_3=[2,4,2,1,5,1,1,2,3,3]$. (The same colors means jointness). \textcolor[rgb]{0,0,0}{(b) Correlation source matrix for the source model shown in (a). Black and white squares mean 1 and 0, respectively, and blue square indicates that its value is unknown (0 or 1).}}
			\label{fig:FL1}
		\end{figure}
		
		The proposed algorithm is applied to analyze simulated and real fMRI datasets either to extract the underlying fMRI components or to discriminate multi-subject datasets that have partially-joint sources and divide them into separate groups. In our simulation study, we will show that the proposed algorithm extracts latent sources of simulated fMRI data and determines the type of sources (joint, partially-joint and individual) with the accuracy of 95\% and 100\% for 2-class and 3-class clustering scenarios, respectively. The algorithm also clusters the datasets that have similar sources with the accuracy higher than 94\% for both 2-class and 3-class clustering scenarios. We also compare our algorithm with the CNFE \cite{ref30} and JI-ThICA \cite{pakravan2018} alternative methods and show the superiority of the proposed algorithm in extracting partially-joint sources.
		
		Finally, we investigate two real fMRI datasets: i) \textit{same story different story} dataset \cite{yeshurun2017} and ii) \textit{Depression} dataset \cite{lepping2016}. Our results in analyzing real datasets reveal that both datasets follow the introduced JpJI-MDM source model. Furthermore, we extract joint and partially-joint sources that are consistent with the existing literature. 
		
		The rest of the paper is organized as follows. Section II is devoted to present the proposed algorithm including signal model, optimization cost function, the method used to determine the type of sources, and comparison of \textcolor[rgb]{0,0,0}{CNFE in \cite{ref30},} JI-ThICA in \cite{pakravan2018}, and the proposed method. In Section III, numerical results for simulated and real fMRI data are reported. Finally, concluding remarks are presented in Section IV.
		
		\subsection{Notation}
		In this paper, matrices are denoted by capital letters, e.g., $Y^{(k)}$ or $M^{(c,k)}$, where the superscripts $(k)$ and $(c,k)$ indicate the index of  ``subject'' and index of  ``source, subject'', respectively. Furthermore, vectors are represented by boldface lowercase letters, e.g., \textcolor[rgb]{0,0,0}{$c$th} raw of the matrix $Y^{(k)}$ is indicated by $\mathbf{y}^{(c,k)}$; and scalars are denoted by lowercase letters, e.g., the $v$th entry of the vector $\mathbf{y}^{(c,k)}$ is denoted by $y^{(c,k)}(v)$. The values of indices are typically ranging from 1 to their capital version, e.g., $k=1,...,K$.

		\section{Proposed Algorithm}
		
		\subsection {Signal Model}
		
		In our JpJI-MDM model, we assume that there are $K$ subjects and the $k$th subject has $C^{(k)}$ latent sources, where $C_1^{(k)}$ out of these sources are considered to be joint with all other subjects; and $C_2^{(k)}$ components are joint with a subset of subjects, which are referred to  as partially-joint sources. Besides, $C_3^{(k)}$ is the number of sources that are independent of all sources in all other subjects (denoted as individual sources).  Note that the total numbers of sources ($C^{(k)}={C_1}^{(k)}+{C_2}^{(k)}+C_3^{(k)}$) in different subjects are not necessarily equal.
		
		Fig. \ref{fig:FL1}(a) shows an example of JpJI-MDM source model, in which $\mathbf{s}^{(c,k)}_{J}$, $\mathbf{s}^{(c,k)}_{PJ}$, and $\mathbf{s}^{(c,k)}_{I}$ indicate that the $c$th source in $k$th subject are  joint source across all subjects, partially-joint source across a subset of subjects, and individual source, respectively. Note that the same colors means jointness.  In Fig. \ref{fig:FL1}(a), we have \textcolor[rgb]{0,0,0}{$C_1=2$}, $C_2=[3,2,3,3,1,3,2,3,3,2]$, and $C_3=[2,4,2,1,5,1,1,2,3,3]$.
		
		It should be noted that our algorithm can analyze any clustering scenarios, in which there are two or more clusters of subjects for partially-joint sources, and subjects can be clustered into either disjoint or no disjoint clusters. Furthermore, it is clear that in JpJI-MDM source model, ${C_1}^{(1)}={C_1}^{(2)}=...={C_1}^{(K)}=C_1$.
		
		We define $\tilde{\mathcal{K}}_{c,k}$ as a subset of \textcolor[rgb]{0,0,0}{subject}s in which the $c$th source is common with the $k$th \textcolor[rgb]{0,0,0}{subject}, and $n_{\tilde{\mathcal{K}}_{c,k}}$ is the number of \textcolor[rgb]{0,0,0}{subject}s in $\tilde{\mathcal{K}}_{c,k}$. According to the JpJI-MDM model we have
		\begin{align}
		\label{eq01}
		{n_{\tilde{\mathcal{K}}_{c,k}}}=\begin{cases}
		K-1 ,& \operatorname{if~}\left(c \in \mathcal{J}\right)\\
		\text{less than $K-1$} ,& \operatorname{if~}\left(c \in \mathcal{PJ}\right)\\
		0 ,& \operatorname{if~}\left(c \in \mathcal{I}\right)\\	
		\end{cases}
		\end{align} 
		where $\mathcal{J}$, $\mathcal{PJ}$, and $\mathcal{I}$ are the set of joint, partially-joint, and individual sources, respectively.
		It should be noted that if $c$th source in $k$th \textcolor[rgb]{0,0,0}{subject} is individual ($c \in \mathcal{I}$), it means that it only is joint with itself, and $n_{\tilde{\mathcal{K}}_{c,k}}=0$.
		
		In Fig. \ref{fig:FL1}(b), the cross-correlation matrices of subjects $k_1$ and $k_2$ are depicted for two scenarios: ($k_1=k_2$) and ($k_1\neq k_2$). For the case $k_1\neq k_2$, there are some blue squares, which their values are 1 if $k_1$ and $k_2$ are in the same cluster; otherwise, they are 0.
		Note that the JpJI-MDM source model can be reduced to SDU, MDU and JI-MDM models if we set ($K$=1), ($K>$1, $C_2^{(k)}$=0 and $C_3^{(k)}$=0 for $k=1,...,K$), and ($K>$1, $C_2^{(k)}$=0 for $k=1,...,K$), respectively. Furthermore, if $K>$1, $C_1$=0, $C_2^{(k)}$=0, $C_3^{(k)}\neq0$ for $k=1,...,K$, the JpJI-MDM model represents $K$ separate and independent SDU source models. 
		
		Let $V$ and $N$ denote the number of voxels in an fMRI image and the number of time samples in an observed blood oxygen level dependent (BOLD) signal, respectively. The vectorized versions of fMRI source images (or volumes) are processed in our algorithm. In addition, $O^{(k)}$ represents the observation signal of $k$th subject, which is related to the mixing matrix ($A^{(k)}\in \mathbb{R}^{N\times C^{(k)}}$) and source matrix ($S^{(k)}$) as follows
		\begin{align}
		\label{eq1}
		O^{(k)}=A^{(k)} S^{(k)},~~~ k=1,...,K.
		\end{align}
		Note that $S^{(k)}=[{\mathbf{s}^{(1,k)}}^T, {\mathbf{s}^{(2,k)}}^{T},..., \mathbf{s}^{(C^{(k)},k)^{T}}]^T$, where $\mathbf{s}^{(c,k)}\in \mathbb{R}^{1\times V}$ for $1\leq c\leq C^{(K)}$ indicates $c$th source. Here, we assume that the elements of $S^{(k)}$ ($\mathbf{s}^{(c,k)}$) are mutually independent and locally stationary. In addition, these elements are normalized to have unit variance and zero mean. The mixing matrices $A^{(k)};k=1,...,K$ are also full-column rank. 
		
		In our study, latent sources are extracted by exploiting cumulants and cross-cumulants. In multiple datasets, $\operatorname{cum}\left[S^{(k_1)},S^{(k_2)}\right]$, $\operatorname{cum}\left[S^{(k_1)},S^{(k_2)},S^{(k_3)}\right]$ and $\operatorname{cum}\left[S^{(k_1)},S^{(k_2)},S^{(k_3)},S^{(k_4)}\right]$ are all tensors in $\mathbb{R}^{C^{(k_1)}\times C^{(k_2)}}$, $\mathbb{R}^{C^{(k_1)}\times C^{(k_2)}\times C^{(k_3)}}$, and $\mathbb{R}^{C^{(k_1)}\times C^{(k_2)}\times C^{(k_3)}\times C^{(k_4)}}$, respectively, which are given by
		{\small
			\begin{align}
			\label{eq4_1}
			\left(\operatorname{cum}\left[S^{(k_1)},S^{(k_2)}\right]\right)_{(c_1,c_2)}= \frac{1}{V}\sum_{v=1}^V \operatorname{cum}\left[s^{(c_1,k_1)}(v),s^{(c_2,k_2)}(v)\right]
			\end{align}
			\begin{align}
			\label{eq4_2}
			&\left(\operatorname{cum}\left[S^{(k_1)},S^{(k_2)},S^{(k_3)}\right]\right)_{(c_1,c_2,c_3)}=\nonumber\\
			&~~~~~~~~~~~~~~~~~~\frac{1}{V}\sum_{v=1}^V \operatorname{cum}\left[s^{(c_1,k_1)}(v),s^{(c_2,k_2)}(v),s^{(c_3,k_3)}(v)\right]
			\end{align}
			\begin{align}
			\label{eq4_3}
			&\left(\operatorname{cum}\left[S^{(k_1)},S^{(k_2)},S^{(k_3)},S^{(k_4)}\right]\right)_{(c_1,c_2,c_3,c_4)}=\nonumber\\
			&~~~~~~~~~~~~~~\frac{1}{V}\sum_{v=1}^V \operatorname{cum}\left[s^{(c_1,k_1)}(v),s^{(c_2,k_2)}(v),s^{(c_3,k_3)}(v),s^{(c_4,k_4)}(v)\right].
			\end{align}}
		Note that $s^{(c,k)}(v)$ denotes the $(c,v)$th element of $S^{(k)}$. In MDM, JI-MDM, and JpJI-MDM source models, nonzero elements are only for $c_1=c_2=c_3=c_4=c$. Furthermore, for the JpJI-MDM model, we have
		\begin{align}
		\label{eq4_4}
		\operatorname{cum}\left[s^{(c,k_1)}(v),...,s^{(c,k_{\eta})}(v)\right]\begin{cases}
		\neq 0~\operatorname{~if~}\left(\forall k_{i_1} \text{\&} k_{i_2}; k_{i_1} \in \tilde{\mathcal{K}}_{c,k_{i_2}}\right) \\
		=0~~\operatorname{~if~}\left(\exists k_{i_1} \textcolor[rgb]{0,0,0}{\text{\&} k_{i_2}}; k_{i_1} \notin \tilde{\mathcal{K}}_{c,k_{i_2}}\right)
		\end{cases}
		\end{align} 
		where $\eta=2,3,4$ and $i_1,i_2=1,...,\textcolor[rgb]{0,0,0}{\eta}$.

		\subsection{Optimization Cost function}
		In the proposed algorithm, a deflation framework \cite{ref1} is employed to extract latent sources across multiple subjects. In this algorithm, a cost function, referred to as joint/partially-joint/individual ICA (JpJI-ICA), is used to extract sources of each subject one-by-one. 
		
		In order to generate the preprocessed observations matrices, $Z^{(k)}$; first, the principle component analysis (PCA) method is applied to reduce the dimension of the observations. The number of principle components $C^{(k)}$ for each subject can be estimated by utilizing model order selection approaches using information-theoretic criteria such as Bayesian information criterion (BIC) \cite{ref7} and Akaike information criterion (AIC) \cite{ref6}. In this study, we employ BIC to estimate $C^{(k)}$ for $k=1,...,K$. Then, a pre-whitening system ($W_w^{(k)}={R_{Z^{(k)}}}^{-1/2}$) is employed \cite{ref1}, where $R_{Z^{(k)}}$ is the covariance matrix of $Z^{(k)}$. It is worth mentioning that in the rest of the paper, we define matrix $U^{(k)}$ as the demixing matrix for $k$th subject ($S^{(k)} = U^{(k)} Z^{(k)}$).
		
		In our algorithm, higher order cumulants are selected as the base of the proposed cost function, because the local convergence of higher order cumulants is not affected with the distributions of sources with non-zero cumulant \cite{ref5}. In order to analyze multi-subject datasets based on the JpJI-MDM model, we need to extract joint, partially-joint, and individual sources of subjects in one algorithm. In this case, we have defined three metrics; the first one is a measure of cross-cumulant between subjects ($C_{(\alpha,\eta)}^{(c,\tilde{\mathcal{K}})}(v)$ in equation \eqref{eq12_1} ), the second one is a sum of weighted norm of cross-cumulants ($D_{(\alpha)}^{(c,k)}(v)$ in equation \eqref{eq12_2}), and the third one computes $D_{(\alpha)}^{(c,k)}(v)$ across subjects that belong to the set $\tilde{\mathcal{K}}$ and all of voxels ($M_{\mathcal{\tilde{K}}}^{(c,k)}$ in equation \eqref{eq12_3}).
		\begin{align}
		\label{eq12_1}	
		C_{(\alpha,\eta)}^{(c,\tilde{\mathcal{K}})}(v)=\operatorname{cum}\left[\mathbf{z}^{(k)}(v),y^{(c,\tilde{\mathcal{K}}(\alpha))}(v),...,y^{(c,\tilde{\mathcal{K}}(\alpha+\eta-2))}(v)\right],
		\end{align}
		\begin{align}
		\label{eq12_2}	
		\textcolor[rgb]{0,0,0}{D_{(\alpha)}^{(c,k)}(v)=\sum_{\eta} w_{\eta} C_{(\alpha,\eta)}^{(c,\tilde{\mathcal{K}})}(v)\left(C_{(\alpha,\eta)}^{(c,\tilde{\mathcal{K}})}(v)\right)^T},
		\end{align}
		and
		\begin{align}
		\label{eq12_3}
		M_{\mathcal{\tilde{K}}}^{(c,k)}=\frac{1}{V}\sum_{v} \sum_{\alpha=1}^{n_{\tilde{\mathcal{K}}}} D_{(\alpha)}^{(c,k)}(v),
		\end{align}
		where $\eta=2,3,4$, ${\mathcal{\tilde{K}}}=\text{randperm}(\{1,...,K\}\setminus \{k\})$, $n_{\tilde{\mathcal{K}}}$ is the size of the set ${\tilde{\mathcal{K}}}$ ($n_{\tilde{\mathcal{K}}}=K-1$), and $\tilde{\mathcal{K}}(\alpha)$ is the $\alpha$th element of the set ${\mathcal{\tilde{K}}}$. Note that if $\alpha+\eta-2>n_{\tilde{\mathcal{K}}}$ for $\eta=3,4$ we choose mod$(\alpha+\eta-2,n_{\tilde{\mathcal{K}}})$ as the new subject index, where mod(,) denotes the modulo operation.
		We also assume that the number of subjects with joint or partially-joint sources are bigger than $\eta-1$ ($n_{\tilde{\mathcal{K}}}>\eta-1$). $w_{\eta}>0$ is the weight considered for the $\eta$th order cumulant; $v=1,...,V$ indicates the index of pixels or voxels; \textcolor[rgb]{0,0,0}{$Y^{(k)}=U^{(k)}Z^{(k)}$} represents the estimated source \textcolor[rgb]{0,0,0}{matrix}, in which $\mathbf{u}^{(c,k)}$ denotes the $c$th raw of the estimated demixing matrix \textcolor[rgb]{0,0,0}{($U^{(k)}$)}; and $\mathbf{z}^{(k)}(v)$ is the $v$th column of $Z^{(k)}$. 
		
		Note that in equation \eqref{eq12_3}, for each  $1\leq \alpha \leq \textcolor[rgb]{0,0,0}{n_{\tilde{\mathcal{K}}}}$ in $\sum_{\alpha}$, we use \{$k$, $\tilde{\mathcal{K}}(\alpha)$,$\tilde{\mathcal{K}}(\alpha+1)$, $\tilde{\mathcal{K}}(\alpha+2)$\}th subjects in cross-cumulants. If  \textcolor[rgb]{0,0,0}{$c$th} sources of these selected subjects are joint (i.e., \textcolor[rgb]{0,0,0}{$\left\{\tilde{\mathcal{K}}(\alpha),\tilde{\mathcal{K}}(\alpha+1),\tilde{\mathcal{K}}(\alpha+2)\right\}\in \tilde{\mathcal{K}}_{c,k}$}), then $M_{\mathcal{\tilde{K}}}^{(c,k)}$ gets its maximum value, \textcolor[rgb]{0,0,0}{equation} \eqref{eq12}.  In other words, for joint sources, the values of $D_{(\alpha)}^{(c,k)}(v)$ for different $\alpha$s are equal (i.e., $D_{(1)}^{(c,k)}(v)=...=D_{(n_{\tilde{\mathcal{K}}})}^{(c,k)}(v)$), hence we can write
		\begin{align}
		\label{eq12}
		M_{\mathcal{\tilde{K}}}^{(c,k)}(v)=(K-1) D_{(1)}^{(c,k)}(v)  \operatorname{~if~}\left(c \in \mathcal{J}\right),
		\end{align}
		On the other hand, in cases that \textcolor[rgb]{0,0,0}{$c$th} source is partially-joint or individual, all four selected subjects do not have necessarily joint sources. This is due to the fact that the indices of subjects are selected randomly, thus we have
		\begin{align}
		\label{eq12e}
		\textcolor[rgb]{0,0,0}{M_{\mathcal{\tilde{K}}}^{(c,k)}(v)< n_{\tilde{\mathcal{K}}_{c,k}} D_{(\alpha_0)}^{(c,k)}(v),}
		\end{align}
		\textcolor[rgb]{0,0,0}{where in $D_{(\alpha_0)}^{(c,k)}(v)$, the indices $\alpha_0$, ..., $\alpha_0+\eta-2$ are the index of $\eta-1$ of subjects that are joint with \textcolor[rgb]{0,0,0}{$k$th} subject in \textcolor[rgb]{0,0,0}{$c$th} source (i.e., $\{\tilde{\mathcal{K}}(\alpha_0),\tilde{\mathcal{K}}(\alpha_0+1),\tilde{\mathcal{K}}(\alpha_0+2)\}$ $\subset \tilde{\mathcal{K}}_{c,k}$}).
		
		By using equations \eqref{eq12_1}, \eqref{eq12_2}, and \eqref{eq12_3}, we define a measure of cross-cumulant between estimated source and sources in other subjects as follows
		\begin{align}
		\label{eq12_t}	
		\Upsilon_{\tilde{\mathcal{K}}} (\mathbf{y}^{(c,k)}) =\mathbf{u}^{(c,k)}\left(M_{\mathcal{\tilde{K}}}^{(c,k)}\right)\mathbf{u}^{(c,k)^T}.
		\end{align}
		Based on equation \eqref{eq12_1}, it is clear that in order to extract desired sources and to align joint sources across subjects in \textcolor[rgb]{0,0,0}{equation} \eqref{eq12_t}, it is not necessary to determine $\tilde{\mathcal{K}}_{c,k}$ for each $c$th source in $k$th subject, because the algorithm automatically converts the cost function from $\tilde{\mathcal{K}}$ to $\tilde{\mathcal{K}}_{c,k}$. This is due to the fact that in \textcolor[rgb]{0,0,0}{equation} \eqref{eq12_1}, only subjects that are members of $\tilde{\mathcal{K}}_{c,k}$, are important. Thereby, if $\mathcal{K}_{c,k}$ has no member, it means that the $c$th source is an individual source of the $k$th subject, and accordingly the cost function will be zero. In this case, we set $\tilde{\mathcal{K}}=\left\{k\right\}$, and use equations \eqref{eq12_1}, \eqref{eq12_2}, \eqref{eq12_3}, and \eqref{eq12_t} to extract individual sources.
		
		By considering Theorem 1 in \cite{pakravan2018} \textcolor[rgb]{0,0,0}{and \cite{ref2}}, it can be shown that a local maximum of the JpJI-ICA cost function (equation \eqref{eq12_t}) can be used to extract one of joint or partially-joint sources. We assume that there is a permutation set $\delta$ of indexes $\left\{1,...,C^{(k)}\right\}$ for independent components of $k$th subject so that the following inequalities hold for $k=1,...,K$.
		\begin{align}
		\label{eq7}
		&\Upsilon_{\tilde{\mathcal{K}}} ({\mathbf{y}^{(\delta_1,k)}}^*)>...>\Upsilon_{\tilde{\mathcal{K}}}({\mathbf{y}^{(\delta_{C^{(k)}},k)}}^*).
		\end{align}
		
		It is worth mentioning that in \textcolor[rgb]{0,0,0}{equation} \eqref{eq7} the relation may not be absolutely increasing and $>$ can be replaced with $\geq$. However, for the absolutely increasing relation we obtain unique ordering for the extracted sources during different runs.
		
		Thanks to the JpJI-ICA cost function, the permutation indeterminacy across subjects is resolved, and dependent sources are automatically grouped across subjects. Furthermore, in this method, parameters $C_1$, $C_2^{(k)}$, and $C_3^{(k)}$ are not necessary to be known for each subject, because the algorithm automatically determines the type of each source. Note that the type of each extracted source (joint, partially-joint, or individual) can be used as another information to cluster joint subjects, which is discussed in subsection II.C.

		The optimization problem in our algorithm is
		\begin{align}
		\label{equ5_6}
		\mathbf{u}^{(c,k)^*} = \underset{\mathbf{u}^{(c,k)}}{\operatorname{argmax}}~\Upsilon_{\tilde{\mathcal{K}}} (\mathbf{y}^{(c,k)})~\text{subject to } |\mathbf{u}^{(c,k)}|^2=1,
		\end{align}
		where ${\mathbf{y}^{(c,k)}}^*= {\mathbf{u}^{(c,k)}}^* Z^{(k)}$ is the desired source, and ${\mathbf{u}^{(c,k)}}^*$ is the optimum value of ${\mathbf{u}^{(c,k)}}$. It is straightforward to show that the optimum value of $\mathbf{u}^{(c,k)}$ is the corresponding eigenvector of the dominant eigenvalue of $M_{\mathcal{\tilde{K}}}^{(c,k)}\in \mathbb{R}^{C^{(k)}\times C^{(k)}}$. The extracting vector $\mathbf{u}^{(c,k)}$ can be determined by \textcolor[rgb]{0,0,0}{thin-SVD factorization}.
		
		\textcolor[rgb]{0,0,0}{In the Theorems 1 and 2 of  \cite{cruces2004thin}, it has been shown that in each iteration, the choice resulting from the thin-SVD factorization guarantees a monotonous ascent in the cost function, and because of the bounded nature of the cost function, the only strictly stable points of the algorithm are the local maxima. Inspired by these Theorems, it is easy to show that the JpJI-ICA cost function has a monotonous ascent which its stable points are the local maxima and do not depend on $\tilde{\mathcal{K}}$s (see Fig. 4 in the supplementary materials).}
		
		\textcolor[rgb]{0,0,0}{On the other hand, in \cite{ref2} it has been shown  that the global maximum of the cost function leads to the extraction of the desired sources. Furthermore, Theorem 2 of this paper showed that although deceptive local maxima of the cost function might exist, but under some assumptions (which are similar to the assumptions considered in the JpJI-ICA cost function), the only local maxima of the cost function correspond to the solutions that extract one of the sources.}
		
\begin{figure}[t!]
			\centering
			\includegraphics[width=3.5in,trim=0cm 0cm 0cm 0cm,clip]{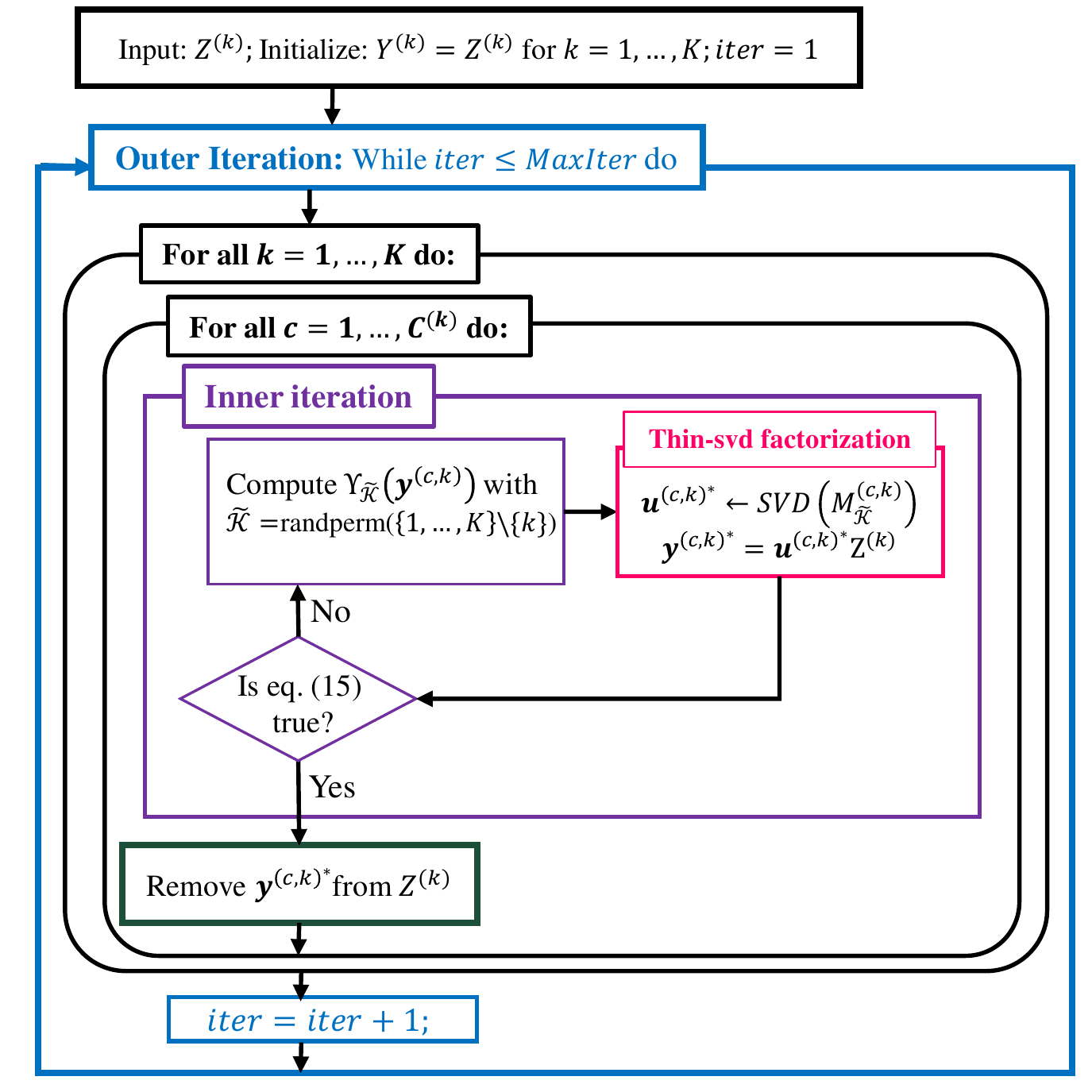}
			\caption{\textcolor[rgb]{0,0,0}{Illustrative flowchart of the JpJI-ICA algorithm.}}
			\label{fig:FLa}
		\end{figure}

		In the proposed algorithm, two iteration loops are considered to maximize the JpJI-ICA cost function, referred to as \textit{inner iteration} and \textit{outer iteration}. The inner iteration is repeated while the following inequality holds for $\epsilon_0=10^{-6}$,
		\begin{align}
		\label{eq1_o}
		1-\left|\mathbf{u}^{(c,k)}{\mathbf{u}^{(c_{old},k)}}^T\right|^2\geq \epsilon_0.
		\end{align}
		Furthermore, the outer iteration is executed $MaxIter$ times to extract all optimum sources of all subjects. Note that in the first iteration of the inner iteration, for each $c$ and $k$ we set its observation as the initial guess of the desired source; and after extracting a desired independent component ($\mathbf{y}^{(c,k)^*} \in \mathbb{R}^{1\times V}$) in the inner iteration, its contribution is subtracted from the observation matrix ($Z^{(k)}$) by linear regression to obtain each row of new observation matrix\cite{ref1}. In Fig. \ref{fig:FLa}, the illustrative flowchart of the proposed algorithm is presented.

		\subsection{Determining the type of sources}
		We define $\Upsilon_{\tilde{\mathcal{K}}} (\mathbf{y}^{(c,k)})$ for $\tilde{\mathcal{K}}=\text{randperm}(\{1,\cdots,K\} \setminus \{k\})$ as \textit{JpJI-feature} for source $c$ of $k$th subject ($JpJIF^{(c,k)}$). Furthermore, we have defined the number of subjects (except \textcolor[rgb]{0,0,0}{$k$th} subject) that have $c$th source (i.e., $n_{\tilde{\mathcal{K}}_{(c,k)}}$) as the jointness of $c$th source in \textcolor[rgb]{0,0,0}{$k$th} subject. Therefore, it can be concluded that the \textit{upper bound} of $JpJIF^{(c,k)}$ has a linear relationship with the amount of its jointness with other subjects (inequality \eqref{eq12e}). Note that for joint sources, the $JpJIF^{(c,k)}$ has a linear relationship with the amount of its jointness with other subjects (equality \eqref{eq12}).
		It is worth mentioning that in \textcolor[rgb]{0,0,0}{equation} \eqref{eq12_t}, $\Upsilon_{\tilde{\mathcal{K}}} (\mathbf{y}^{(c,k)})$ is zero (or very low) for subjects in which $c$th source is neither joint or partially-joint, because the cross-cumulant of  mutually-statistically-independent random variables in \textcolor[rgb]{0,0,0}{equation} \eqref{eq12_t} is zero. Therefore, the following relations are always true:
		\begin{align}
		\label{equ5_16}
		&JpJIF^{(c,k)}\text{ if $c\in \mathcal{J}$}>JpJIF^{(c,k)}\text{ if $c\in \mathcal{PJ}$}>JpJIF^{(c,k)}\text{ if $c\in \mathcal{I}$}.
		\end{align}
		
		\textcolor[rgb]{0,0,0}{In order to determine the type of extracted sources, two different approaches are proposed based on using either the JpJI-feature or the spatial sources. In the first approach, the $JpJIF^{(c,k)}$ is used to determine the type of extracted $c$th source in the $k$th subject, as follows:}		
		\begin{align}
		\label{eq12_f}
		\begin{cases}
		\textcolor[rgb]{0,0,0}{\text{If~}JpJIF^{(c,k)} \approx (K-1)\frac{1}{V}\sum_{v} D_{(1)}^{(c,k)}(v) \rightarrow \left(c \in \mathcal{J}\right)}\\
		\textcolor[rgb]{0,0,0}{\text{If~}\sigma_0<JpJIF^{(c,k)}\neq (K-1)\frac{1}{V}\sum_{v} D_{(\alpha_0)}^{(c,k)}(v) \rightarrow \left(c \in \mathcal{PJ}\right)}\\
		\textcolor[rgb]{0,0,0}{\text{If~}JpJIF^{(c,k)}\leq \sigma_0 \rightarrow \left(c \in \mathcal{I}\right)}\\
		\end{cases}
		\end{align}

		\textcolor[rgb]{0,0,0}{In this approach, first, we determine the joint sources, then we decide about the type of the rest of sources using $\sigma_0$. To find the optimum threshold for $\sigma_0$ ($\sigma_{opt}$) without using the label of sources, we have applied a simple and efficient method based on the following steps:}
		\begin{enumerate}
			\item \textcolor[rgb]{0,0,0}{Define the mean value of JpJI-features for joint sources as $JpJIF^{Joint}$.}
			\item \textcolor[rgb]{0,0,0}{Compute the parameter  $Ratio(c)=\frac{JpJIF^{Joint}}{\text{mean}(JpJIF^{(c,:)})}$ for $C_1< c\leq C^{(k)};1\leq k \leq K$. Since the JpJI-feature of partially-joint sources are closer to $JpJIF^{Joint}$ than the  individual ones, then the $Ratio(c)$ of individual sources are very bigger than the partially-joint ones (e.g., $Ratio(c)$ for partially-joint and individual sources are approximately in the range of [1 , $10^2$] and [$10^6$, $10^8$], respectively).}
			\item \textcolor[rgb]{0,0,0}{Cluster $Ratio$ into 2 clusters (i.e., partially-joint cluster and individual cluster) using \textit{kmeans} algorithm.}
			\item  \textcolor[rgb]{0,0,0}{Determine the lower bound and higher bound of $\sigma_{opt}$ as $\text{max}(JpJIF^{(c,:)})$ for  $c \in\mathcal{I}$ and $\text{min}(JpJIF^{(c,:)})$ for $c \in\mathcal{PJ}$, respectively.}
			\item  \textcolor[rgb]{0,0,0}{$\sigma_{opt}=\frac{\text{lower bound}+\text{higher bound}}{2}$.}
		\end{enumerate}
		
		\textcolor[rgb]{0,0,0}{Simulation results show that using $Ratio$ has better performance than JpJI-features, because the $Ratio$ of partially-joint and individual sources have more different values than their JpJI-features. It is worth mentioning that if there are no joint sources, then one can use the gap between the JpJI-features of partially-joint and individual sources for clustering.}

		\textcolor[rgb]{0,0,0}{In the second approach, the spatial shape of the extracted sources are used to determine their type. Since the JpJI-ICA algorithm groups joint sources in a similar source index, we have used the following method to determine the type of $c$th source.}
		\begin{enumerate}
			\item \textcolor[rgb]{0,0,0}{Group the $c$th source of all subjects in two classes (i.e., random grouping or grouping in predefined classes).}
			\item \textcolor[rgb]{0,0,0}{Define a statistical test on each voxel with this null hypothesis that the values of sources in one group of subjects are similar to other subjects' group with a predefined FDR corrected $P_{value}$.}
			\item \textcolor[rgb]{0,0,0}{If the null hypothesis is rejected and some significant cluster of voxels can be found that their activities are different in two groups of the subjects, then the $c$th source is partially-joint or individual; otherwise it is a joint source.}
			\item \textcolor[rgb]{0,0,0}{To discriminate between partially-joint and individual sources, one can redefine a similar statistical test on the different subgroups of subjects in each group which is made in step 1. If the null hypothesis is rejected, the $c$th source is partially-joint, otherwise, it is an individual source.}
		\end{enumerate}
		
		\textcolor[rgb]{0,0,0}{It is worth mentioning that if the goal is classification or clustering of the subjects based on their partially-joint sources in different classes, then using the ``JpJI-feature" approach is more straightforward (as a feature for discrimination). On the other hand, if the goal is to find significant ROIs in the brains of subjects that had different activities in some predefined classes of subjects, then it is better to use the ``spatial source approach", because it gives automatically significant cluster of voxels in which their activity are significantly different across groups of subjects.}\\

		After determining the type of each source, one can simply determine $\tilde{\mathcal{K}}_{c,k}$ for each $c$ and $k$. Note that for joint and individual sources we have $\tilde{\mathcal{K}}_{c,k}=\tilde{\mathcal{K}}$ and $\tilde{\mathcal{K}}_{c,k}=\{\}$, respectively. In addition, \textit{k-means} clustering algorithm \cite{macqueen1967some} is applied to automatically cluster subjects of partially-joint sources and determine $\tilde{\mathcal{K}}_{c,k}$ for $k=1,...,K$ and $c\in\mathcal{PJ}$.

		\subsection{Comparing JpJI-ICA with JI-ThICA \cite{pakravan2018}}
		As mentioned, in \cite{pakravan2018} the JI-MDM source model, has been assumed and the so-called JI-ThICA algorithm has been proposed to extract latent sources. There are some similarities and differences between JI-ThICA and JpJI-ICA algorithms. Both methods extract the latent sources by maximizing the sum of weighted higher order cross-cumulants among subjects. However, their procedures to determine the type of sources are different. The JI-ThICA algorithm introduces two cost functions to extract joint and individual sources separately, but the JpJI-ICA algorithm introduces one unique cost function for joint, partially joint, and individual sources. The JI-ThICA method uses the following cost functions to determine the type of sources:		
		\begin{align}
		\label{equThin2}
		\textcolor[rgb]{0,0,0}{\Upsilon(\mathbf{y}^{(c,k)})=\mathbf{u}^{(c,k)}\left(\tilde{\mathcal{M}}^{(c,k)}\right)\mathbf{u}^{(c,k)^T}, \text{~for } c \in \mathcal{J};}
		\end{align}
		\begin{align}
		\label{equThin3}
		\Psi(\mathbf{y}^{(c,k)})=\mathbf{u}^{(c,k)}\left(M^{(c,k)}\right)\mathbf{u}^{(c,k)^T}, \text{~for } c \in \mathcal{I};
		\end{align}
		where $\tilde{\mathcal{M}}^{(c,k)}$ and $M^{(c,k)}$ are computed as follows
		\begin{align}
		\label{eqThin1_1}
		\tilde{\mathcal{M}}^{(c,k)}=\frac{1}{V}\sum_{v} \sum_{\eta =2,3,4}w_{\eta}{\left|cum\left[\mathbf{z}^{(k)}(v),y^{(c,k_2)}(v),...,y^{(c,k_\eta)}(v)\right]\right|}^2,
		\end{align}
		\begin{align}
		\label{eqThin1_2}
		M^{(c,k)}=\frac{1}{V}\sum_{v} \sum_{\eta=2,3,4}w_{\eta}{\left|cum\left[\mathbf{z}^{(k)}(v),y^{(c,k)}(v),...,y^{(c,k)}(v)\right]\right|}^2,
		\end{align}
		where in \textcolor[rgb]{0,0,0}{equation} \eqref{eqThin1_1} $k_j\in \{1,...,K\}\setminus \{k\}$ for $j=2,...,\eta$ indicate index of subjects. In addition, the operator $\left|.\right|^2$ denotes the squared norm. Note that these $\eta-1$ subjects are selected randomly from $\{1,...,K\}\setminus \{k\}$, which are indexed as $k_2$,..., $k_\eta$ subjects. If  the value of $\Upsilon(\mathbf{y}^{(c,k)})$ is higher than a given threshold ($\sigma_0$), then this method assumes that all of these selected subjects are joint with \textcolor[rgb]{0,0,0}{$k$th} subject, and the JI-ThICA algorithm determines $c$th source as joint and maximizes \textcolor[rgb]{0,0,0}{equation} \eqref{equThin2} to extract this source. On the other hand, if the value of $\Upsilon(\mathbf{y}^{(c,k)})$ is lower than $\sigma_0$, then the JI-ThICA method assumes that the source is independent from other subjects, and it determines $c$th source as individual and  maximizes \textcolor[rgb]{0,0,0}{equation} \eqref{equThin3} to extract this source. This means that JI-ThICA cannot discriminate partially-joint sources and considers partially-joint sources randomly as joint or individual sources.
		
		In the JpJI-ICA algorithm, we have combined \textcolor[rgb]{0,0,0}{equation} \eqref{equThin2} and \textcolor[rgb]{0,0,0}{equation} \eqref{equThin3}  in one cost function by introducing a new parameter ($\tilde{\mathcal{K}}$)  where ${\mathcal{\tilde{K}}}=\text{randperm}(\{1,...,K\}\setminus \{k\})$ as follows
		\begin{align}
		\label{equThin4}
		&\Upsilon_{\tilde{\mathcal{K}}} (\mathbf{y}^{(c,k)})=\mathbf{u}^{(c,k)}\left(M_{\mathcal{\tilde{K}}}^{(c,k)}\right)\mathbf{u}^{(c,k)^T}
		\end{align}
		in which $M_{\mathcal{\tilde{K}}}^{(c,k)}$ is computed as follows
		\begin{align}
		\label{eqThin5}
		M_{\mathcal{\tilde{K}}}^{(c,k)}=&\frac{1}{V}\sum_{v} \sum_{\alpha=1}^{n_{\tilde{\mathcal{K}}}} \sum_{\eta=2,3,4} w_{\eta} \nonumber\\	
		&\left|\operatorname{cum}\left[\mathbf{z}^{(k)}(v),y^{(c,\tilde{\mathcal{K}}(\alpha))}(v),...,y^{(c,\tilde{\mathcal{K}}(\alpha+\eta-2))}(v)\right]\right|^2
		\end{align}
		The main difference of two methods is the summation $\sum_{\alpha=1}^{n_{\tilde{\mathcal{K}}}}$ in \textcolor[rgb]{0,0,0}{equation} \eqref{eqThin5} which enables JpJI-ICA algorithm to discriminate partially-joint sources.  Accordingly, in the JpJI-ICA algorithm, the equation \eqref{eq12_f} is used to determine the type of sources.

		\section{Numerical Results}
		\subsection{Evaluation Metrics}
		The joint Signal-to-Interference Ratio (jSIR) \cite{ref30} is computed to evaluate performance of the proposed algorithm as $jSIR = \frac{1}{K}\sum_{k} \frac{1}{C^{(k)}}\sum_c 10 \text{log}_{10}⁡\left(\frac{\left(\mathbf{s}^{(c,k)}\right)\left({\mathbf{y}^{(c,k)}}^*\right)^T}{\left(\mathbf{s}^{(c,k)}-{\mathbf{y}^{(c,k)}}^*\right)\left(\mathbf{s}^{(c,k)}-{\mathbf{y}^{(c,k)}}^*\right)^T}\right)$, where ${\mathbf{y}^{(c,k)}}^*$ and $\mathbf{s}^{(c,k)}$ denote normalized estimated and real sources, respectively. Note that ${\mathbf{y}^{(c,k)}}^*$ and $\mathbf{s}^{(c,k)}$ have zero mean and unit variance. The higher jSIR means lower error between the real and estimated sources, which is desired in our performance evaluation. It is worth mentioning that the fraction inside the logarithm function indicates the ratio of correlation and distance of two vectors, thus, jSIR is the difference of correlation and distance of two vectors in dB. More details on the comparison of the jSIR and correlation metrics are presented in Fig. 1 in the supplementary material.
		
		The accuracy of JpJI-ICA algorithm in determining the correct numbers of joint, partially-joint, and individual sources is evaluated by $Acc(\mathcal{C})$, which is computed as $Acc(\mathcal{C})=\frac{1}{N_{Run}}\sum_{r} Acc(\mathcal{C},{r})$ where $Acc(\mathcal{C},{r})$ denote the correctness of the estimated number of sources (separately for joint, partially-joint, and individual) in $r$th run of the algorithm, and $N_{Run}$ is the total number of runs. Note that $Acc(\mathcal{C},{r})$ is a Boolean variable, and it is 100 if the estimated number of sources is correct; otherwise, it is 0.

		We also evaluate the accuracy of the proposed algorithm in clustering subjects with respect to their partially-joint sources by computing $Acc(\tilde{\mathcal{K}}_{all})=\frac{1}{N_{Run}}\sum_{r} \frac{1}{K}\sum_{k} \frac{1}{C^{(k)}}\sum_c Acc(\tilde{\mathcal{K}}_{c,k},r)$, where $Acc(\tilde{\mathcal{K}}_{c,k},r)$ is a Boolean variable denoting the accuracy of the algorithm in its $r$th run to find $\tilde{\mathcal{K}}_{c,k}$ for $k$th subject. If all elements of the estimated $\tilde{\mathcal{K}}_{c,k}$ are correct, then $Acc(\tilde{\mathcal{K}}_{c,k},r)=100$; otherwise $Acc(\tilde{\mathcal{K}}_{c,k},r)=0$.
		
		\subsection{Results on Simulated fMRI Data}
		We synthesize multi-subject fMRI datasets by using the SimTB toolbox\footnote{This toolbox is available at http://mialab.mrn.org/software/simtb/ [Accessed: 2017-03-20].} \cite{ref17} to evaluate the JpJI-ICA algorithm. The SimTB toolbox in its default settings generates 27 spatially independent brain sources, and in this study, we use these pre-generated sources. We generate fMRI images with $64\times 64$ pixels. Furthermore, the same synthesized spatial maps are assigned across all subjects for joint sources, whereas different spatial maps are assigned for individual sources. Moreover, time courses generated with SimTB toolbox are used as a mixing matrix. Interested readers are referred to see \cite{ref17} for further details about the SimTB toolbox and simulated fMRI datasets. Spatial ICA is applied to analyze fMRI data in which a weighted linear combination of $C^{(k)}$ spatial sources compose each fMRI image of $k$th subject over time.
		
		It should be noted that the number of simulated sources in SimTB toolbox is limited to 27, thus, there are some limitations to increase $K$, $C_1$, $C_2^{(k)}$, and $C_3^{(k)}$ in the simulation of JpJI-MDM source model. Fig. 2(a-b) in the supplementary material show an example of a simulated JpJI-MDM source model with SimTB toolbox and  the estimated sources with JpJI-ICA algorithm with jSIR=21.3 dB and correct estimations for source types, respectively.
		
		Though the JpJI-ICA cost function should be optimized with respect to the weights $w_2$, $w_3$, and $w_4$, here for the sake of brevity, we set $w_2=0.5$, $w_3=0.75$, $w_4=1$. Note that in \cite{pakravan2018}, different weights settings for order cumulants in the JI-MDM source model have been analyzed, and it has been shown that the fourth order cumulant has the most contribution on the performance improvement, and using the second and third order cumulants can slightly improve the performance of the JI-ThICA algorithm. Our results in this study reveal that this conclusion is also valid for the JpJI-ICA cost function. 
		
		The number of latent sources of each subject ($C^{(k)}$) is estimated by utilizing the BIC method with the maximum likelihood ICA algorithm \cite{ref10}. \textcolor[rgb]{0,0,0}{Fig. 3 in the supplementary material presents the impact of the estimated model order for simulated data on the results of the JpJI-ICA algorithm.}
		
		We have designed some experiments to evaluate the performance of JpJI-ICA algorithm in terms of its convergence rate and its accuracy to extract correct latent sources (in terms of mean jSIR), estimation of the correct numbers of joint, partially-joint, and individual sources, and determining $\tilde{\mathcal{K}}_{c,k}$ for all sources of all subjects. 
		
\begin{figure}[t!]
			\centering
			\includegraphics[width=3.5in,trim=0cm 0cm 0cm 0cm,clip]{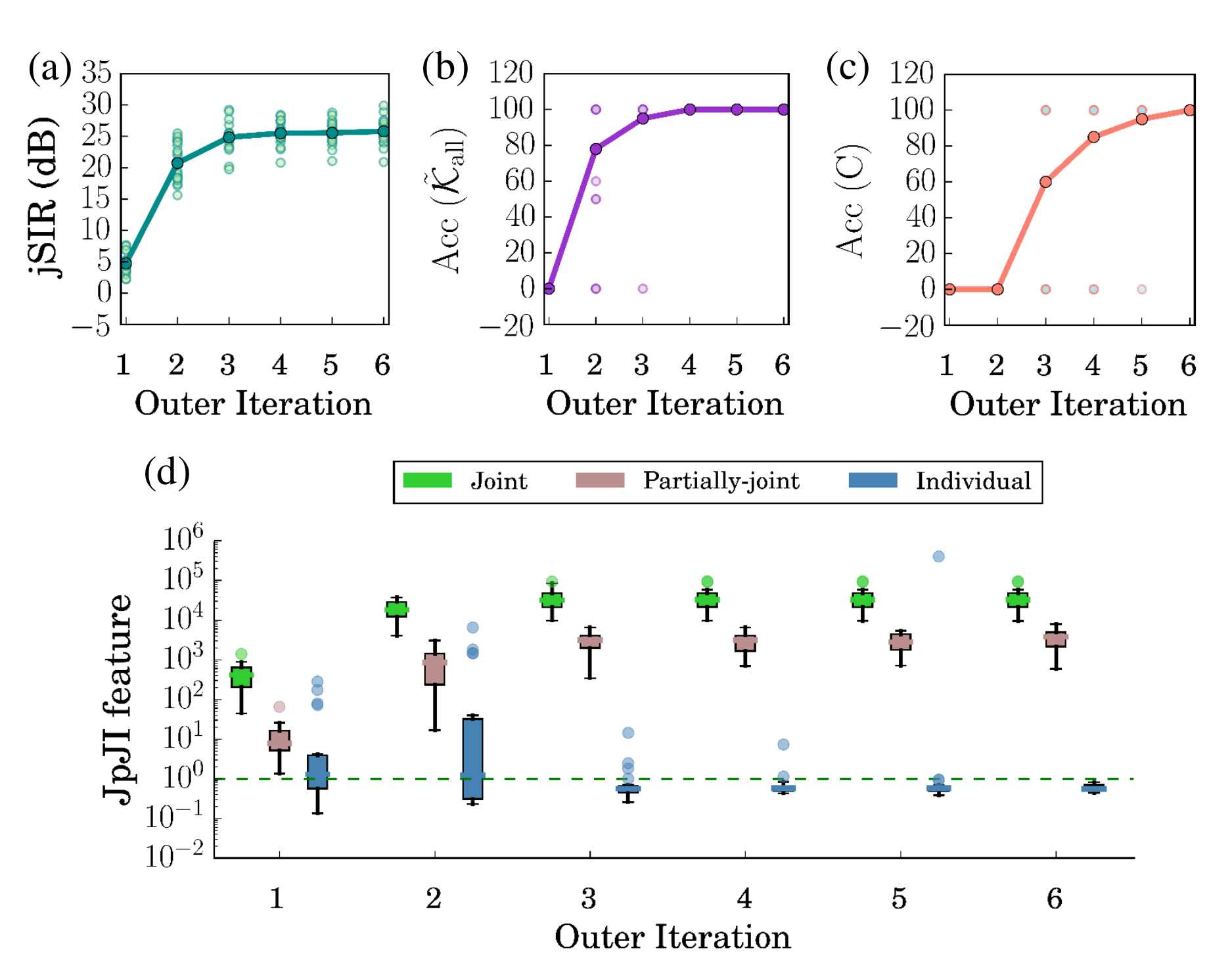}
			\caption{(a-d) Performance of the JpJI-ICA algorithm versus different iterations in terms of (a) the convergence rate of jSIR (dB), (b) $Acc(C)$, (c) $Acc(\tilde{\mathcal{K}}_{all})$, (d) JpJI-feature for $K=10, C_1=3,C_2^{(k)}=2,C_3^{(k)}=1,k=1,...,K$ versus outer iterations.}
			\label{fig:FL04}
		\end{figure}

		We have analyzed the convergence rate of the algorithm to evaluate its performance when the outer iterations increases. Figs. \ref{fig:FL04}(a-c) show the convergence rate of the JpJI-ICA algorithm in terms of jSIR (dB), $Acc(C)$, and $Acc(\tilde{\mathcal{K}}_{all})$ for $K=10, C_1=3,C_2^{(k)}=2,C_3^{(k)}=1,k=1,...,K$ versus outer iterations. Fig. \ref{fig:FL04}(d) represents the JpJI-feature separately for joint, partially-joint, and individual sources versus outer iteration. The results are obtained from 50 Monte Carlo runs. Results reveal that 5 or 6 iterations of the outer iteration are sufficient for the convergence of the algorithm. \textcolor[rgb]{0,0,0}{ In the rest of the paper, we set $MaxIter=5$}.
		
		\begin{figure}[t!]
			\centering
			\includegraphics[width=3.5in,trim=0cm 0cm 0cm 0cm,clip]{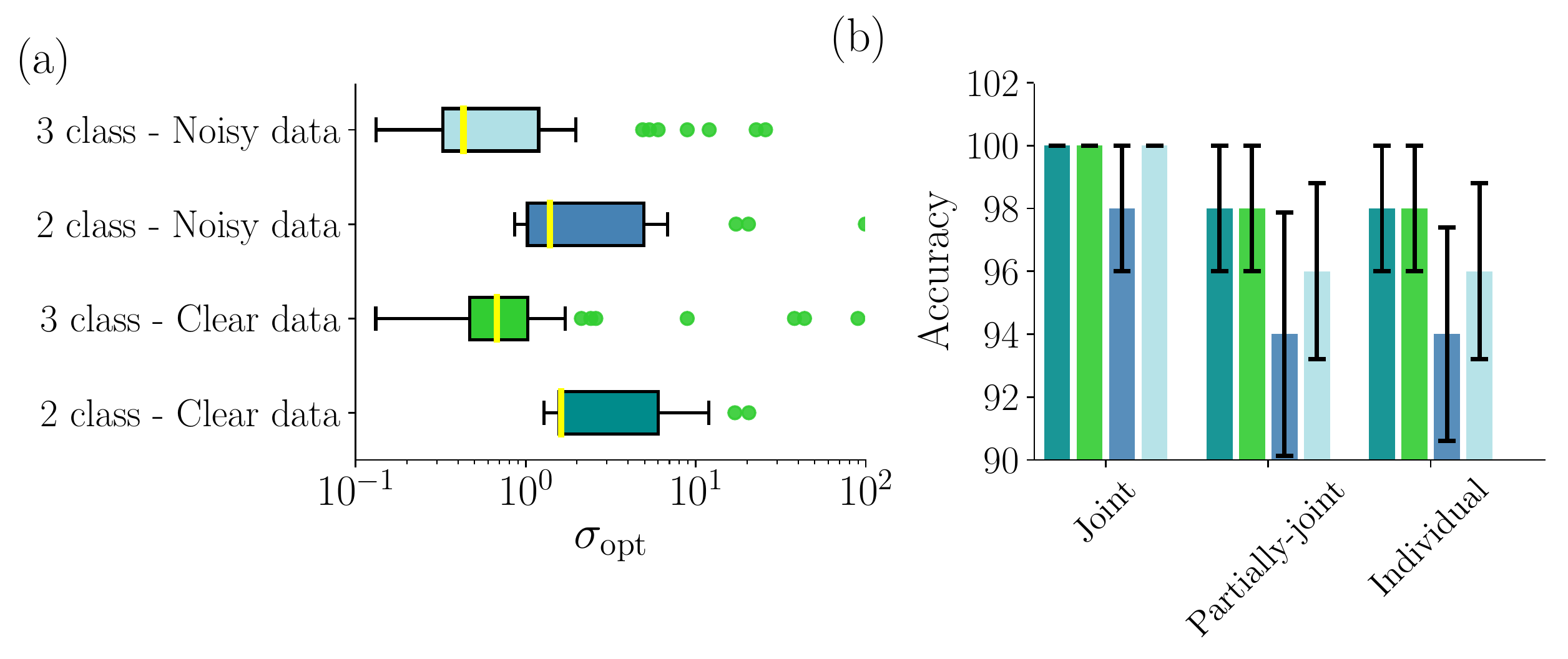}
			\caption{\textcolor[rgb]{0,0,0}{Performance of the JpJI-ICA algorithm in determining optimum value for $\sigma_0$ ($\sigma_{opt}$) in the different 50 Monte-Carlo runs ($K=10$, $C_1$ and $C_2^{(k)}$ are selected randomly, $C_3^{(k)}=1,k=1,...,K$) for 4 different simulation scenarios, (a) optimum range for $\sigma_0$, (b) accuracy of determining the type of each joint, partially-joint and individual sources separately.}}
			\label{fig:FL04new}
		\end{figure}

		We have analyzed the performance of the algorithm when $\sigma_0$ changes manually. Performance of JpJI-ICA algorithm in determining the type of each extracted source is presented in the Fig. 5 in the supplementary material.
		\textcolor[rgb]{0,0,0}{Fig. \ref{fig:FL04new}(a) presents the range of obtained $\sigma_{opt}$ for four different simulation scenarios and 50 Monte-Carlo runs  ($K=10$, $C_1$ and $C_2^{(k)}$ are selected randomly, and $C_3^{(k)}=1, k=1,...,K$). After determining $\sigma_{opt}$, the accuracy of estimating correct number of joint, partially-joint, and individual sources are shown in Fig. \ref{fig:FL04new}(b) separately.}
		
		\textcolor[rgb]{0,0,0}{It is worth mentioning that for the simulated fMRI datasets, we apply JpJI-feature based approach to determine the type of sources and select $\sigma_{opt}$ as the threshold to discriminate the partially-joint and individual sources. However, for real fMRI datasets, we use source based approach to determine the type of each extracted sources and obtain significant clusters of voxels that are different in some groups of subjects based on their partially-joint sources.}
		
		Furthermore, we have prepared two scenarios to evaluate the performance of the algorithm. In the first scenario, we assume that there are only two clusters of subjects that have similar partially-joint sources. The results of this scenario are referred to as 2-class. Similarly, in the second scenario, we assume that there are three clusters of subjects that have similar partially-joint sources. The results of this scenario are referred to as 3-class. We also compare results of clear and noisy observations, where for noisy observations, a white Gaussian noise is added to the observations with signal to noise ratio (SNR) of 3dB, where $SNR = 10 \left(log_{10} \frac{Signal\ Power}{Noise\ Power}\right)$. 
		
\begin{figure*}[t!]
			\centering
			\includegraphics[width=6in,trim=0cm 0cm 0cm 0cm,clip]{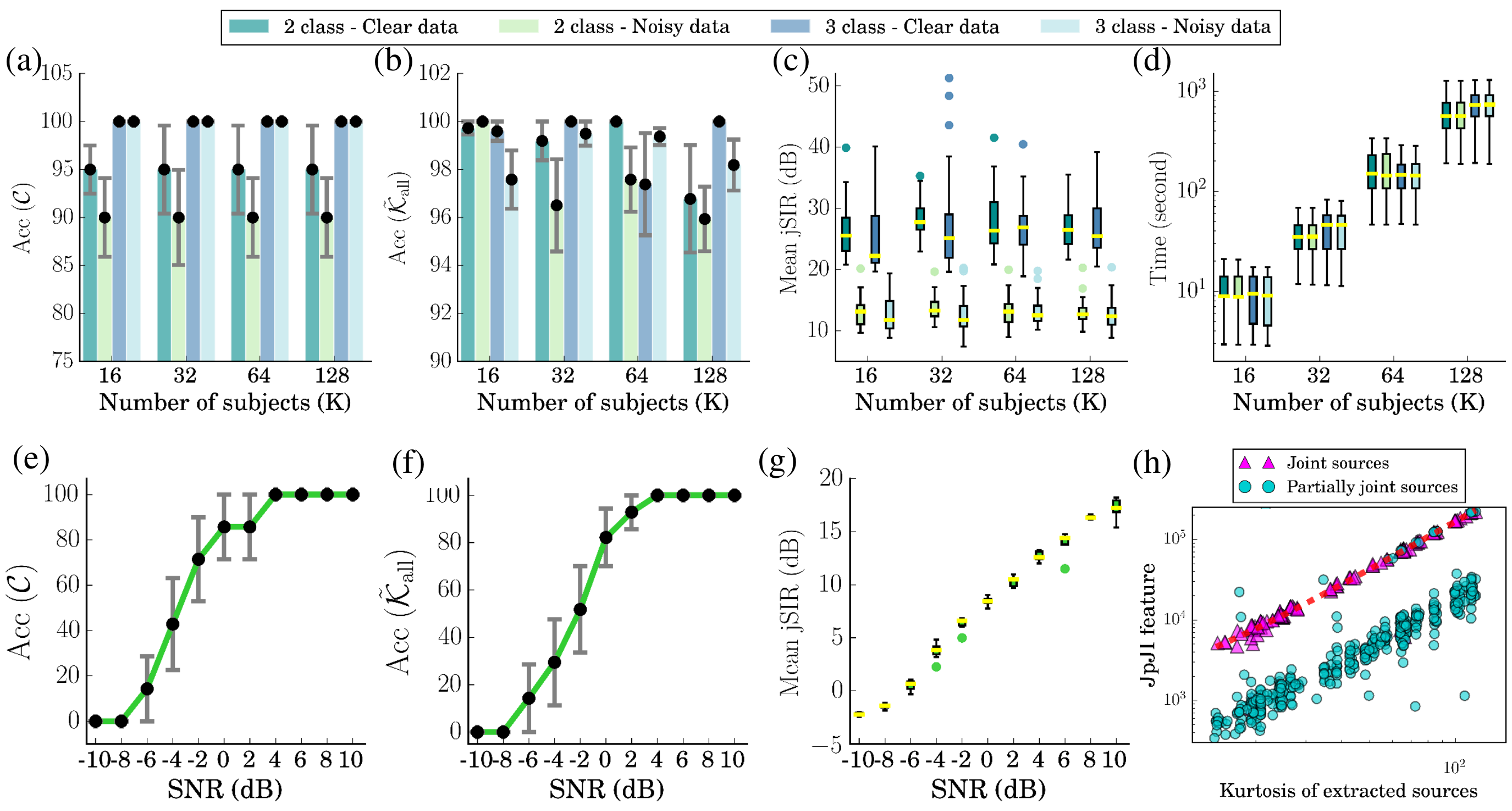}
			\caption{(a-d) Performance of the JpJI-ICA algorithm versus the number of subjects, (a) accuracy of estimating correct number of sources for clear and noisy observations (with SNR 3db), (b) mean accuracy of determining $\tilde{\mathcal{K}}_{c,k}$ for all $c$ and $k$ based on the partially-joint sources, (c) mean jSIR (dB), and (d) mean run time in seconds; (e-g) performance of  the JpJI-ICA algorithm versus the SNR of observations in dB ($K=10,C_3^{(k)}=1,k=1,...,K$, $C_1$ and $C_2^{(k)}$ are selected randomly) in terms of (e) its accuracy to estimate the number of joint and partially-joint sources, (f) its accuracy to determine $\tilde{\mathcal{K}}_{c,k}$ for all $c$ and $k$ for partially-joint sources, (g) mean jSIR (dB); and (h) relating JpJI-feature and the Kurtosis of extracted source ($K=16,C_1=3,C_2^{(k)}=2,C_3^{(k)}=0,k=1,...,K$).}
			\label{fig:FL05}
		\end{figure*}

		In Figs. \ref{fig:FL05}(a-d), simulation results are shown for different numbers of subjects ($K$). Note that the results are obtained from 50 Monte Carlo runs in which $C_1$ and $C_2^{(k)}$ are selected randomly to simulate source data. As mentioned, since the number of simulated sources in SimTB toolbox is limited to 27, only for $K=16$, the numbers of individual sources are nonzero (i.e., $N = 150$, $0\leq C_1,C_2^{(k)}\leq 8$, $C_1$+ $C_2^{(k)}\neq 0$, $C_3^{(k)}=1$ for $k=1,...,K$ and $K=16$). However, for $K>16$, subjects do not have individual sources (i.e., $C_3^{(k)}=0$ for $k=1,...,K$ and $K>16$).
		
		Figs. \ref{fig:FL05}(a) and \ref{fig:FL05}(b) show accuracy of our algorithm to estimate the number of joint, partially-joint, and individual sources, $Acc(\mathcal{C})$, and to cluster subjects with their partially-joint sources, $Acc(\tilde{\mathcal{K}}_{all})$, for clear and noisy observations, respectively. In \ref{fig:FL05}(a), we observe that the algorithm has a better $Acc(\mathcal{C})$ in the 3-class scenario, and its clustering accuracy is approximately the same in all cases. In addition, the algorithm also clusters the subjects that have similar sources with accuracy higher than 94\% in both 2 and 3 class clustering scenarios (Fig \ref{fig:FL05}(b)).
		
		In Figs. \ref{fig:FL05}(c) and \ref{fig:FL05}(d), the mean jSIR and run time of the algorithm are depicted for clear and noisy observations, respectively. As expected, clear observations have a better jSIR, and the run time of the algorithm is increased by increasing $K$. It should be noted that an Intel(R) Core(TM) i7- 2.40 GHz computer with 8.0 GB of RAM is used to conduct all experiments.
		
		Figs. \ref{fig:FL05}(e-g) represent the performance of the JpJI-ICA algorithm versus the SNR of observations in dB ($K=10,C_3^{(k)}=0,k=1,...,K$, $C_1$ and $C_2^{(k)}$ are selected randomly). We observe that the increase of SNR   improves the accuracy of estimating the number of joint and partially-joint sources (Fig. \ref{fig:FL05}(e)), the accuracy of determining  $\tilde{\mathcal{K}}_{c,k}$ for all $c$ and $k$ for partially-joint sources (Fig. \ref{fig:FL05}(f)), and mean jSIR (Fig. \ref{fig:FL05}(g)).
		
		We have designed further experiments to investigate the relation of JpJI-feature with the spatial shape and type of extracted sources. In the investigated scenarios, the fourth order cumulant has more effect on the JpJI-feature, thus, here we only test the fourth order cumulant of the extracted source as a feature for spatial shape. Note that the forth order cumulant of $\mathbf{y}^{(c,k)}$ with zero mean and unit variance equals to the \textit{kurtosis} of $\mathbf{y}^{(c,k)}$ \cite{ref1}. In Fig. \ref{fig:FL05}(h), the scatter plot of the JpJI-feature versus the kurtosis of extracted sources are depicted ($C_1=3,C_2^{(k)}=2,C_3^{(k)}=0,k=1,...,16$) for 100 Monte Carlo runs. By fitting a curve on the scatter plot, we found that the JpJI-feature and kurtosis have a linear relation (in logarithmic space) for joint sources, which is depicted with dashed red line (Fig. \ref{fig:FL05}(h)), where $JpJIF^{(c,k)}\approx14.8\left(Kurt\left[\mathbf{y}^{(c,k)}\right]\right)^2+8712.7$.
		
\begin{figure}[t!]
			\centering
			\includegraphics[width=3in]{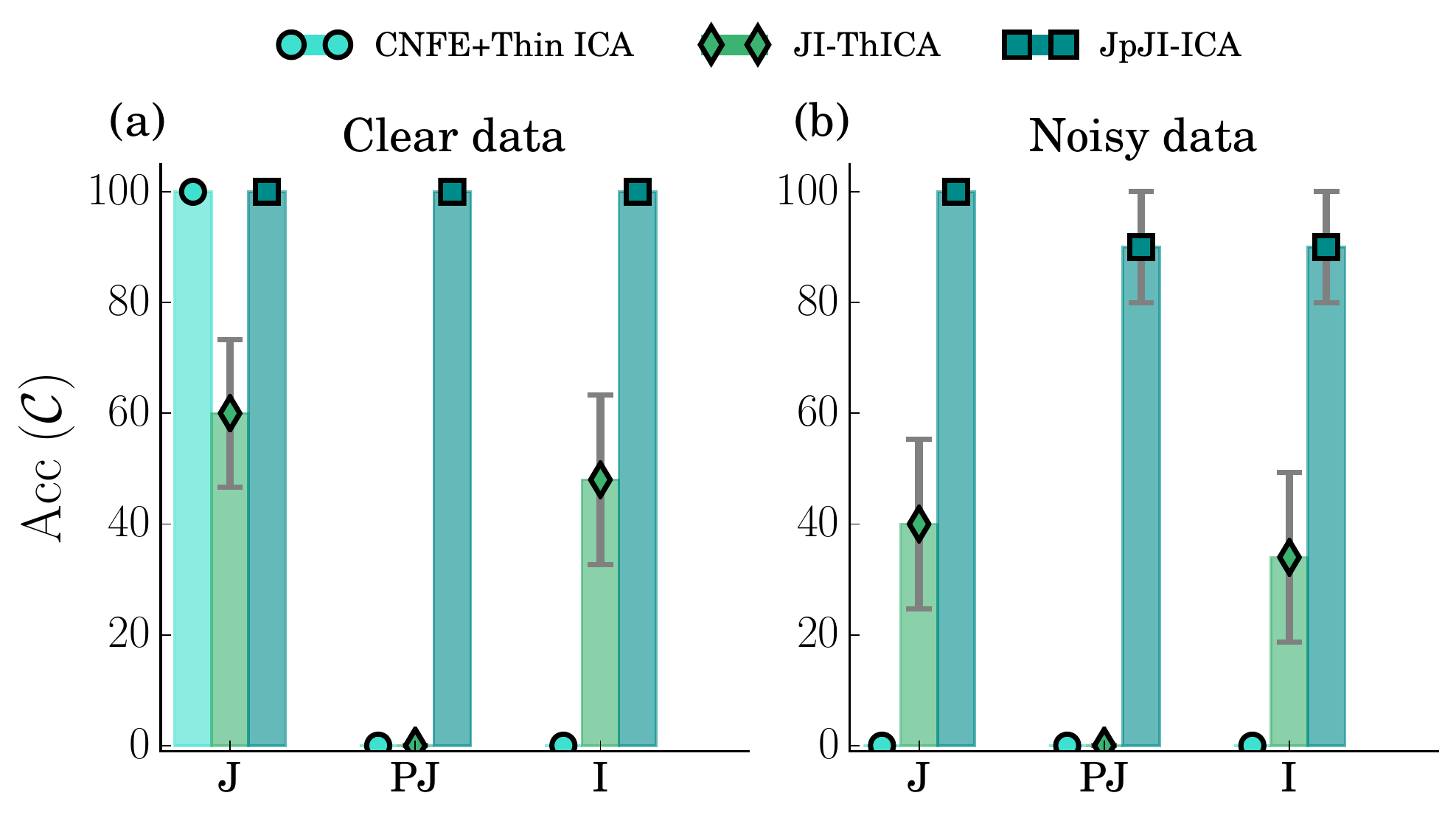}
			\caption{Performance comparison of CNFE+Thin ICA, JI-ThICA, and JpJI-ICA algorithms ($K=10, C_1=2,C_2^{(k)}=2,C_3^{(k)}=1,k=1,...,K$) in terms of their accuracy to estimate the number of joint, partially-joint, and individual sources separately, (a) for clean observations, (b) for noisy observations, where J, PJ, and I mean joint, partially-joint and individual sources, respectively.}
			\label{fig:All}
		\end{figure}

		Although there is no other algorithms with the JpJI-MDM source model, it is interesting to compare the introduced JpJI-ICA method with the existing algorithms designed based on the JI-MDM source model to show the advantage of the JpJI-MDM model over the JI-MDM alternative. To this aim we compare the results of JpJI-ICA, JI-ThICA \cite{pakravan2018}, and CNFE \cite{ref30}, where in CNFE the Thin ICA method \cite{ref2} is used in estimating the number of joint, partially-joint and individual sources  ($K=10, C_1=2,C_2^{(k)}=2,C_3^{(k)}=1,k=1,...,K$). Figs. \ref{fig:All}(a) and \ref{fig:All}(b) show the $Acc(\mathcal{C})$ for clear and noisy observations, respectively. We observe that the CNFE+Thin ICA can estimate the number of joint sources for clear observations, but it cannot discriminate the partially-joint and individual sources (Fig. \ref{fig:All}(a)); on the other hand, this algorithm fails in estimating the number of joint, partially-joint, and individual sources for noisy observations (Fig. \ref{fig:All}b). 
		
		Furthermore, the results in Fig. \ref{fig:All} reveal that the JI-ThICA method cannot discriminate partially-joint sources, and it considers partially-joint sources randomly as joint or individual sources; note that these results confirm our discussion in Section II.D. Thus, the JpJI-ICA algorithm is proposed as a promising solution which determines correctly the numbers of three types of sources (with accuracy of $100\%$ for joint sources in clear and noisy observations, $100\%$ for partially-joint and individual sources in clear observations,  and $95\%\pm5\%$ for partially-joint and individual sources in noisy observations).
		
		Fig. 6 in the supplementary material shows an example for the JpJI-MDM source model. We have also presented the extracted sources by CNFE+thin ICA algorithm, JI-ThICA algorithm, and JpJI-ICA algorithm for noisy observations.
		
		\subsection{Results on Real fMRI Datasets}
		In this section, we introduce the analyzed two real fMRI datasets and present the extracted results.
		
		\subsubsection{Same story, different story dataset \cite{yeshurun2017}}
		
		The interpretation of people about events depends on the external inputs from the world and internal cognitive processes in their brain \cite{anderson1977}. In \cite{yeshurun2017}, the external inputs as stimuli are constant, and the beliefs of two groups of subjects with respect to stimuli are changed before presenting the stimuli. The authors wanted to figure out how the brain conducts the same stimuli in different ways. The dataset is downloaded from the dataspace of Princeton Neuroscience Institute\footnote{This dataset is available at\\ http://arks.princeton.edu/ark:/88435/dsp0141687k93v [Accessed: 2018-10-12]}. 
		
		\textbf{fMRI stimuli}: Forty subjects listened to a short story ``Pretty Mouth and Green My Eyes”. In the story, Arthur has returned home after a party but he could not find his wife, Joanie. He is calling his friend (Lee) to ask him where Joanie is. To disambiguate the story, two different introductions were provided for subjects before listening to the story, \textit{cheating condition} and \textit{paranoia condition}. In the cheating class, the introduction specified that Joanie is cheating on him with Lee. Furthermore, the introduction of the paranoia class mentioned that Arthur is paranoid and his wife is not cheating on him. Subjects were divided in two groups: In the first group, twenty subjects received the first introduction (the story is about a wife cheating on her husband); in the second group, twenty subjects received the second introduction (the story is about a husband being paranoid).
		
		\textbf{fMRI data analysis}: In \cite{yeshurun2017}, fMRI data were reconstructed and analyzed with the BrainVoyager QX software package \cite{goebel2006analysis}. Preprocessing of functional scans included motion correction, slice-time correction, linear-trend removal, and high-pass filtering. Spatial smoothing was applied using a Gaussian filter of 6 $mm$ full-width at half-maximum. The functional data set was transformed to 3-D Talairach space. 
		In \cite{yeshurun2017}, Euclidean distance between the time courses of the two conditions has been computed voxel-by-voxel across the whole brain to test the differences of neural responses of two classes. They also simulated a null distribution by using a permutation method to find significant distance values. They analyzed the whole-brain (voxel-by-voxel level) and predefined set of regions of interest (ROIs). Their predefined ROIs were right and left temporoparietal junction (TPJ), precuneus, posterior cingulate cortex, ventromedial prefrontal cortex (vmPFC), dorsomedial prefrontal cortex (dmPFC), right middle frontal gyrus (MFG), right superior temporal sulcus (STS), right temporal pole, right and left inferior frontal gyrus (IFG), left superior temporal gyrus (STG), and right hippocampus.

		\begin{figure*}[t!]
			\centering
			\includegraphics[width=7in]{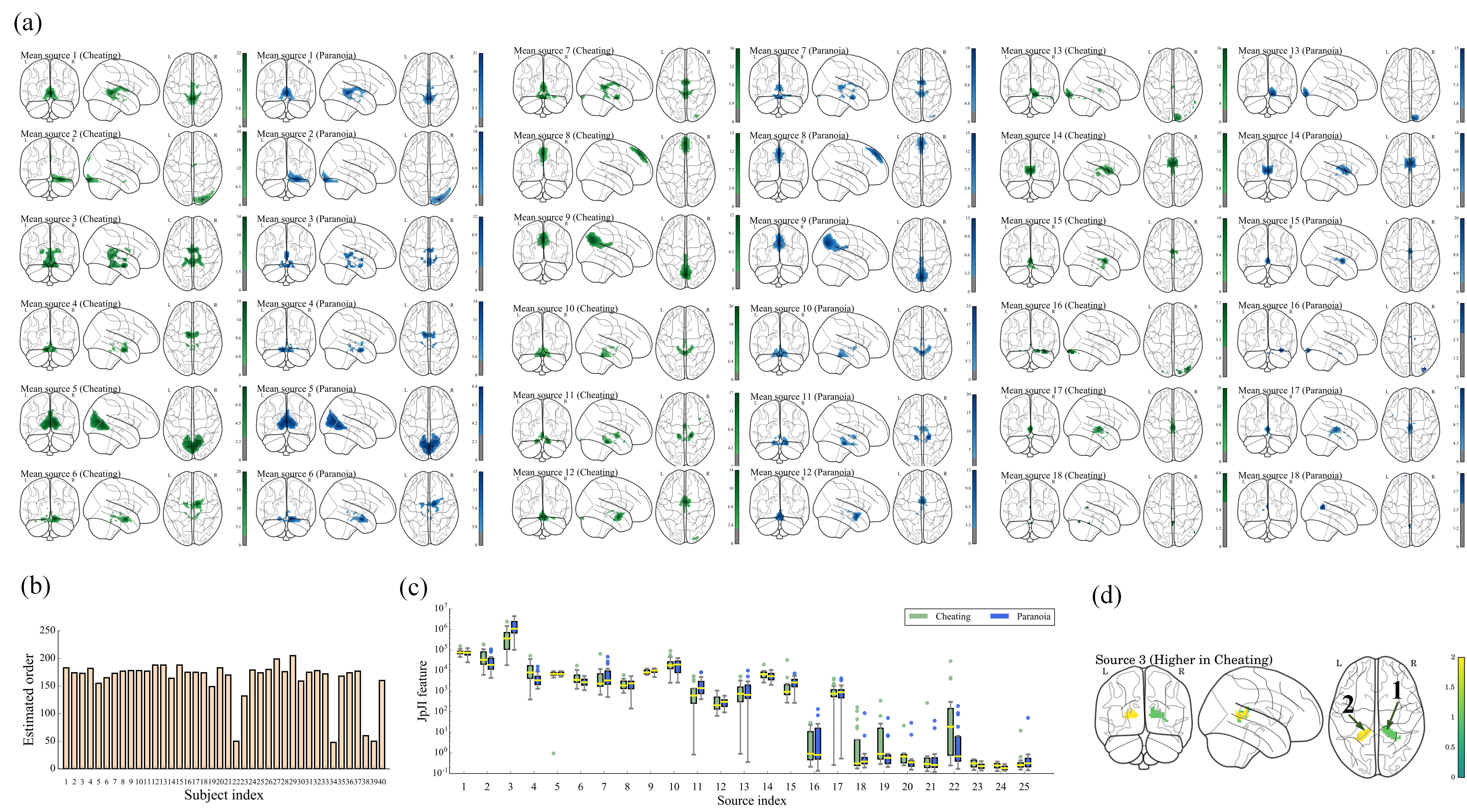}
			\caption{(a) Mean spatial distribution of the first 18 significant sources ($p < 0.05$ FDR corrected) from 20 subjects in the cheating class and 20 subjects in the paranoia class using the JpJI-ICA algorithm, (b) estimated order for 40 subjects using ER-FM model ($C^{(k)},k=1,...,40$) ,(c) JpJI-feature for the first 25 extracted sources, and (d) map of significant clusters in the source 3.}
			\label{fig:GES}
		\end{figure*}
		
		In the current study, we apply the entropy rate (ER)-based order estimation method using finite memory length model (ER-FM) \cite{fu2014} to estimate $C^{(k)};k=1,...,K$ (Fig. \ref{fig:GES}(b)). This order selection model is shown to be a good match of fMRI data \cite{fu2014}. \textcolor[rgb]{0,0,0}{Note that the original number of BOLD signal's time points is 475.}
		
		\textcolor[rgb]{0,0,0}{Since the subjects are from more than one class (e.g., cheating and paranoia), partially-joint sources are more important to find the differences between the brain activities of subjects. On the other hand, it is clear that $\text{min}(C_1+C_2^{(k)}~;~k=1,...,K)\leq \text{min}(C^{(k)}~;~k=1,...,K)$, therefore $\text{min}(C^{(k)}~;~k=1,...,K)$ is considered as the number of independent components for all subjects in order to make sure that at least $\text{min}(C_2^{(k)}~;~k=1,...,K)$ partially-joint sources are extracted.} Then we apply the JpJI-ICA algorithm on the preprocessed fMRI data of 40 subjects to extract 48 independent components.
		
		Fig. \ref{fig:GES}(a) show the mean spatial distribution of significant sources ($p < 0.05$ FDR corrected) of the first 18 extracted independent components for 2 classes of subjects, 20 subjects in the cheating class and 20 subjects in paranoia class.
		\textcolor[rgb]{0,0,0}{It is worth mentioning that the spatial distribution of brain activities in Figs. \ref{fig:GES} and \ref{fig:DSS} are plotted using \textit{nilearn} toolbox \cite{abraham2014machine}.}
		
		As it can be seen in Fig. \ref{fig:GES}(a) the sources with index bigger than 18 for 2 classes are not presented, because their mean spatial map across subjects in each class had no significant voxel with $p < 0.05$ FDR corrected. This is because these extracted sources are individual and their averages across subjects are not significant. Fig. 7 in the supplementary material represents an example of individual sources in sources with the index 23 and 30 for 5 subjects selected randomly (3 from cheating class and 2 from paranoia class). 
		
		Furthermore, Fig. \ref{fig:GES}(c) represents the JpJI-feature for the first 25 extracted sources. This plot shows that for sources with indexes bigger than 22 we can surely conclude that there are no joint or partially-joint sources across cheating and paranoia classes.
		
		Comparisons between the two groups of subjects were made by using \textit{t-tests} in the spatial map of sources, with this null hypothesis that for each voxel, the mean of source values in paranoia condition is equal to the mean of source values in cheating condition ($P_{value}=0.01$). The null hypothesis is rejected only in source 3 (colored clusters in Fig. \ref{fig:GES}(d)). Therefore, using the spatial shapes of extracted sources, we observed that the source 3 is a partially-joint source with disjoint clusters for cheating and paranoia classes. 
		
		In Fig. \ref{fig:GES}(d) the clusters in each map are numbered, where number 1 is related to right Caudate with cluster size of 90 and center of the mass (CM) of cluster with $[+40.7,+10.6,-33.0 ]$ in millimeter; and number 2 is related to left Caudate with cluster size of 68 and cluster CM with $[+43.5,+9.3,+24.0]$. The bilateral Caudate nucleus is significant brain region in which its activity is different in cheating and paranoia condition (bilateral Caudate nucleus is activated in a greater extent in cheating condition), as shown in Fig. \ref{fig:GES}(d). However, in \cite{yeshurun2017}, the Caudate nucleus was not found in the whole brain exploratory analysis as a region that is significantly different in the two conditions. 
		
		It is worth mentioning that the Caudate has been implicated in a variety of cognitive processes \cite{grahn2008cognitive}. For example, tasks that require critical processes for goal-directed actions in a social context robustly activate the Caudate \cite{montague2006imaging}. The role of the Caudate in the complex interactions between social influences and reward is also highlighted by examining the modulation of Caudate activity by perceptions of moral character and altruistic punishment \cite{delgado2005perceptions}. It has also been shown that activation in the human Caudate nucleus is modulated as a function of trial-and-error feedback. For example, in a study, by using a repeated-interaction ``trust'' game, activity in the Caudate decreased over trials as feedback from partners became more predictable \cite{king2005getting}. We can conclude that the Caudate nucleus processes feedback information \cite{delgado2000tracking} especially when feedback is behaviorally relevant \cite{elliott2004nstrumental}. Similarly, in our study, bilateral Caudate nucleus is a significant brain region that behaves differently in mentalizing cheating and paranoia condition. Thus, previous beliefs of two groups of subjects concerning stimuli (as feedback information) influence the activation of Caudate. This finding is consistence with previous studies \cite{elliott2004nstrumental,delgado2000tracking}.
		
		\subsubsection{Depression dataset \cite{lepping2016}}
		
		The goal of this dataset is to compare the brain activity of depressed and control (healthy) subjects when they listen to emotional auditory stimuli, where emotional kinds of music (positive and negative) and sounds (positive and negative) stimuli are used. This data was obtained from the Openneuro data repository. Its accession number is ds000171\footnote{This dataset is available at\\ https://openneuro.org/datasets/ds000171/versions/00001 [Accessed: 2018-10-01].}.
		
		\textbf{fMRI stimuli}: 19 major depressive disorder and 20 never-depressed control subjects listened to positive and negative music and sound stimuli during fMRI scanning. 
		
		\textbf{fMRI data analysis}: 
		In \cite{lepping2016}, statistical parametric methods in the AFNI software were used to analyze activation maps and multiple regression analysis with the general linear model (GLM) was applied to conduct statistical contrasts. They focused on brain regions for emotion processing including anterior cingulate cortex (ACC) (Brodmann areas (BA) 32 and 33, subgenual anterior cingulate cortex (sgACC), and ventral anterior cingulate cortex (vACC)), and striatum (amygdala, Caudate, nucleus accumbens, and putamen). The authors also had whole-brain exploratory analysis.
		
		In this study, we have performed fMRI preprocessing steps by using the AFNI for raw data of 39 subjects including: (1) registering functional images to the first image (using align\_epi\_anat.py), (2) normalizing each participant’s anatomical image to the Montreal Neurological Institute (MNI) template (using align\_epi\_anat.py), (3) registering head motion-corrected functional images to participant’s normalized anatomical image (using align\_epi\_anat.py), (4) slice-timing correction by using heptic interpolation (using 3dTshift function, (5) smoothing by using a Gaussian filter with 6 mm kernel (using 3dmerge function), and (6) high-pass filtering and spike removing the BOLD signals (using 3dBandpass function with fbot=0.009). In addition, the data of 39 subjects for four different stimuli are concatenated and supposed as 156 ($39\times4$) different subjects.
		
\begin{figure*}[t!]
			\centering
			\includegraphics[width=7in]{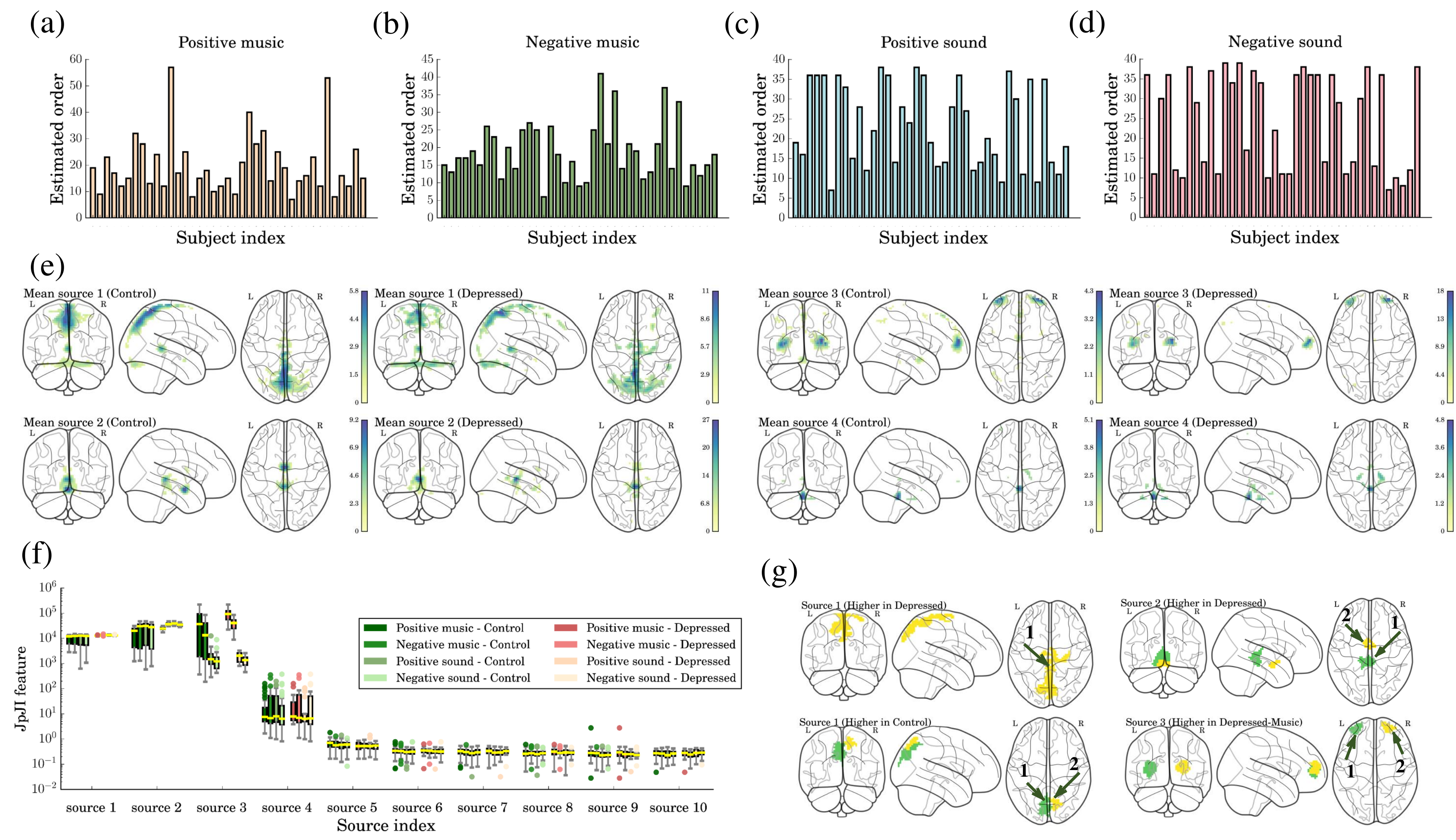}
			\caption{(a-d) Estimated order for 39 subjects ($C^{(k)},k=1,...,39$) with different stimuli types (Positive music, Negative music, Positive sound, Negative sound) using ER-FM model, (e) mean spatial distribution of the first 4 significant sources ($p < 0.05$ FDR corrected) from control and depressed subjects, (f) JpJI-feature for the first 10 extracted sources separately for different stimuli, (g) map of significant voxels in source 1, 2, and 3.}
			\label{fig:DSS}
		\end{figure*}

		We have used the ER-FM order selection model to estimate $C^{(k)};k=1,...,K$ for subjects in different stimuli types (Positive music in Fig. \ref{fig:DSS}(a), Negative music in Fig. \ref{fig:DSS}(b), Positive sound in Fig. \ref{fig:DSS}(c), and Negative sound in Fig. \ref{fig:DSS}(d)). \textcolor[rgb]{0,0,0}{Note that the original number of BOLD signal's time points for music and sound stimuli are 60 and 40, respectively.}

		We have selected $C^{(k)}=10$, $k=1,...,156$  ($\text{min}\left\{C^{(1)},...,C^{(K)}\right\}=7$) because joint sources and partially-joint sources that cluster subjects to two classes (control and depressed) are more important in analyzing the dataset. Then we apply the JpJI-ICA algorithm on the preprocessed fMRI data. 
		Fig. \ref{fig:DSS}(e) shows the mean spatial distribution of significant sources ($p < 0.05$ FDR corrected) of the first 4 extracted independent components for 2 classes of subjects for all 4 stimuli types. The results for depressed and control subjects are presented separately, 20 subjects in the control class and 19 subjects in the depressed class. The sources with index bigger than 5 for 2 classes are not presented, because their mean spatial map had no significant voxel. This is because these extracted sources are individual and their averages across subjects are not significant. Fig. 8 in the supplementary material represents an example of individual sources in the $7$th extracted source for one random subject from the control class and one random subject from the depressed class. These sources are presented for all stimuli types, separately.
		
		\begin{table}[htb]
			\caption{The number, size, center of mass and name of significant clusters in extracted partially-joint sources for "depression" dataset.}
			\resizebox{\columnwidth}{!}{%
				\begin{tabular}{c|ccl|}
					\cline{2-4}
					\multicolumn{1}{l|}{} & \begin{tabular}[c]{@{}c@{}}Cluster\\ size\end{tabular} & CM {[}x,y,z{]} & Cluster name \\ \hline
					\multicolumn{1}{|c|}{\multirow{2}{*}{\begin{tabular}[c]{@{}c@{}}Source 2\\ (Higher in Depressed)\end{tabular}}} & 467 & +3.2 +32.8 -3.4 & (1) left Culmen \\
					\multicolumn{1}{|c|}{} & 253 & +0.7 +3.1 -14.2 & (2) left Subcallosal Gyrus \\ \hline
					\multicolumn{1}{|c|}{\multirow{2}{*}{\begin{tabular}[c]{@{}c@{}}Source 3\\ (Higher in Depressed-Music)\end{tabular}}} & 186 & +31.5 -57.0 
					9.8 & (1) Left Middle Frontal Gyrus \\
					\multicolumn{1}{|c|}{} & 159 & -23.5 -59.0 +12.5 & (2) Right Middle Frontal Gyrus \\ \hline
				\end{tabular}
			}
		\end{table}
		
		Fig. \ref{fig:DSS}(f) represents the mean JpJI-feature of extracted sources for 4 stimuli types and 2 group of subjects, separately.
		
		Fig. \ref{fig:DSS}(g) shows maps of significant clusters (FDR corrected $P_{value}=10^{-2}$) of the first three extracted sources. We observe that in source 1 two different parts of Precuneus are activated for control and depressed subjects. Besides, sources 2 and 3 are higher in depressed subjects with all types of stimuli and depressed subjects with music stimuli, respectively. The clusters in each map are numbered (Fig. \ref{fig:DSS}(g)), and the number, size, center of mass, name, and function of significant clusters in extracted partially-joint sources are listed in Table 2. 
		
		It is worth mentioning that in source 2 the left Culmen has higher activity in depressed patients which is consistent with \cite{su2014cerebral} which comprising 188 depressed patients and 169 healthy controls. Other neuroimaging studies of treatment in depressed patients also show consistent activations in the left Culmen \cite{boccia2016treatment}. On the other hand, the left Subcallosal Gyrus (SCG) has higher activity in depressed subjects in source 2. Our results are consistent with the previous imaging studies which show an increased SCG activity in patients with depression \cite{hamani2011subcallosal,mayberg2005deep,dougherty2003cerebral}. This finding further supports the argument that the SCG is an important region in the pathophysiology of depression. It is worth mentioning that Culmen and SCG were not found as significant brain regions in the original paper \cite{lepping2016}.
		
		In source 3, bilateral Middle Frontal Gyrus (bMFG) has higher activity in depressed patients with music stimuli. In \cite{lepping2016}, it has been shown that control subjects with musical stimuli have a significant activation in bMFG. However, our results show a higher activation in bMFG for depressed subjects with musical stimuli, which are consistent with the existing studies, for example, an increased activation in bMFG to positive stimuli in depressed subjects with respect to control subjects is detected in \cite{canli2004brain,keedwell2005double,lawrence2004subcortical} (happy and sad words \cite{canli2004brain}, happy and sad memory prompts and facial expression \cite{keedwell2005double}, and happy and sad facial expressions \cite{lawrence2004subcortical}).
		
		\section{Conclusions}
		In this paper, we introduced a new source model, referred to as JpJI-MDM, for joint analysis of multi-subject datasets. Accordingly, we presented a new algorithm (JpJI-ICA) to extract three types of sources: joint (common among all subjects), partially-joint (common among a subset of subjects), and individual (specific for each subject).  In the JpJI-ICA algorithm, a deflation framework is employed to extract latent sources across multiple subjects, and sources of each subject are extracted one-by-one by maximizing the cost function of the algorithm. Higher order cumulants are selected as the base of the cost function of the JpJI-ICA algorithm because the local convergence of higher order cumulants is not affected by the distributions of sources with non-zero cumulant. The results from both simulated and real fMRI data showed the benefits of the proposed algorithm as either a complementary or alternative method for the inference of fMRI group data. Furthermore, the JpJI-ICA algorithm can provide a useful interpretation of multi-subject fMRI data and help to deal with more complex and realistic source models. The algorithm determines the types of sources in 2-class and 3-class simulation scenarios with accuracy of 95\% and 100\%, respectively. Also, in both scenarios, subjects with similar sources are grouped with the accuracy higher than 94\%. We also compared our algorithm with its alternatives designed for the JI-MDM source model (the so-called JI-ThICA and CNFE) and showed that it significantly outperforms the existing methods in extracting partially-joint sources.
		By using real fMRI data, we have demonstrated that multi-subject datasets follow the JpJI-MDM source model, and the JpJI-ICA algorithm can extract plausible significant joint and partially-joint spatial maps across different groups of subjects.


	\end{document}